\documentclass{article}

\PassOptionsToPackage{numbers, compress}{natbib}

 \usepackage[preprint]{neurips_2026}

\usepackage[utf8]{inputenc} 
\usepackage[T1]{fontenc}    
\usepackage{hyperref}       
\usepackage{url}            
\usepackage{booktabs}       
\usepackage{amsfonts}       
\usepackage{nicefrac}       
\usepackage{microtype}      
\usepackage{xcolor}         

\usepackage{graphicx}
\usepackage{subcaption}
\usepackage{amsmath}
\usepackage{amssymb}
\usepackage{mathtools}
\usepackage{amsthm}
\usepackage{colortbl}
\usepackage{multicol}
\usepackage{multirow}
\usepackage{soul}
\usepackage{listings}
\usepackage{minted}
\usepackage{caption}
\usepackage{ifthen}
\usepackage{algorithm}
\usepackage{algorithmic}
\usepackage[capitalize,noabbrev]{cleveref}
\usepackage[textsize=tiny]{todonotes}

\graphicspath{{figures/}}

\theoremstyle{plain}
\newtheorem{theorem}{Theorem}[section]

\newtheorem{lemma}[theorem]{Lemma}

\theoremstyle{definition}

\theoremstyle{remark}

\definecolor{lightblue}{RGB}{220,235,255} 
\newcommand{\bbsdp}{\textbf{B}$_{b}$\textbf{-SDP}} 
\newcommand{\bsdp}{\textbf{B-SDP}}
\newcommand{\nsdp}{\textbf{N-SDP}}
\newcommand{\bsdpr}{B-SDP}
\newcommand{\nsdpr}{N-SDP}
\newcommand{\bbsdpr}{B$_{b}$-SDP}
\newcommand{\cor}{$\mathcal{C}$}

\newcommand{\hlb}[1]{{\setlength{\fboxsep}{1pt}\colorbox{lightblue}{#1}}}

\newboolean{useratio}
\setboolean{useratio}{true}

\title{Model Parallelism With Subnetwork Data Parallelism}

\author{\parbox{\textwidth}{\centering
\vspace{0.75cm}
    Vaibhav Singh$^{1,2}$  \hspace{-10pt}
    \qquad Zafir Khalid$^{1,2}$  \hspace{-10pt}
    \qquad Pietro Cagnasso$^{1,2}$ \vspace{5pt}\\
    Edouard Oyallon$^{3}$ \hspace{-10pt}
    \qquad Eugene Belilovsky$^{1, 2}$ \vspace{5pt}\\
    \textnormal{{ $^1$Mila  $^2$Concordia University $^3$ISIR-Sorbonne University, CNRS}}
}}

\makeatletter
\def\blfootnote{\xdef\@thefnmark{}\@footnotetext}
\makeatother

\begin{document}

\maketitle
\blfootnote{Correspondence to: eugene.belilovsky@concordia.ca, vaibhav.singh@mila.quebec, zafir.khalid@mila.quebec}

\begin{abstract}
Pre-training large neural networks at scale imposes heavy memory demands on accelerators and often requires costly communication. We introduce \textbf{Subnetwork Data Parallelism (SDP)}, a distributed training framework that partitions a model into structured subnetworks trained across workers without exchanging activations. We study two complementary masking regimes: \textit{backward masking}, which applies sparsity only in the backward step to retain unbiased gradients, and \textit{forward masking}, which also removes parameters in the forward pass to deliver stronger efficiency gains while providing additional regularization. We further explore two subnetwork construction strategies: \textit{neuron level} and \textit{block level}, applied across both transformers and CNNs.  In experiments spanning 1B LLaMA pre-training on FineWeb to ResNet-18 on CIFAR, SDP reduces per device memory usage by \hlb{\textbf{28\%-60\%}} while maintaining or improving performance under FLOP-matched settings.
\end{abstract}

\section{Introduction}
\label{intro}
The rapid scaling of deep neural networks has led to unprecedented progress across a wide range of domains, from computer vision~\citep{he2016deep, clip, dinov2, segment_anything, shang2024theia} to NLP~\citep{foundationmodels, Achiam2023GPT4TR, touvron2023llama, zhao2023survey}. Training such large models has necessitated distributed strategies like \emph{data parallelism}~\citep{li2020pytorch} and \emph{model parallelism}~\citep{shazeer2018mesh, shoeybi2019megatron, huang2019gpipe}, each with trade-offs. Data parallelism, typically implemented as Distributed Data Parallel (DDP)~\citep{li2020pytorch}, replicates the model on each GPU and synchronizes gradients after every backward pass. While simple and widely used, it incurs high memory overhead from full replication and high communication cost during synchronization. When models are too large to fit on a single accelerator, Model Parallelism (e.g., GPipe~\citep{huang2019gpipe}) mitigates memory issues by splitting layers across devices but requires expensive high-bandwidth interconnects to communicate activations. Unlike DDP, where several methods reduce communication cost~\citep{Diloco,wang2023cocktailsgd}, lowering activation bandwidth remains an open challenge. Moreover, pipeline approaches often suffer inefficiencies from idle waiting (pipeline bubbles).

\begin{figure}[t]
  \centering
  \begin{subfigure}[t]{0.49\linewidth}
    \centering
    \includegraphics[width=\linewidth]{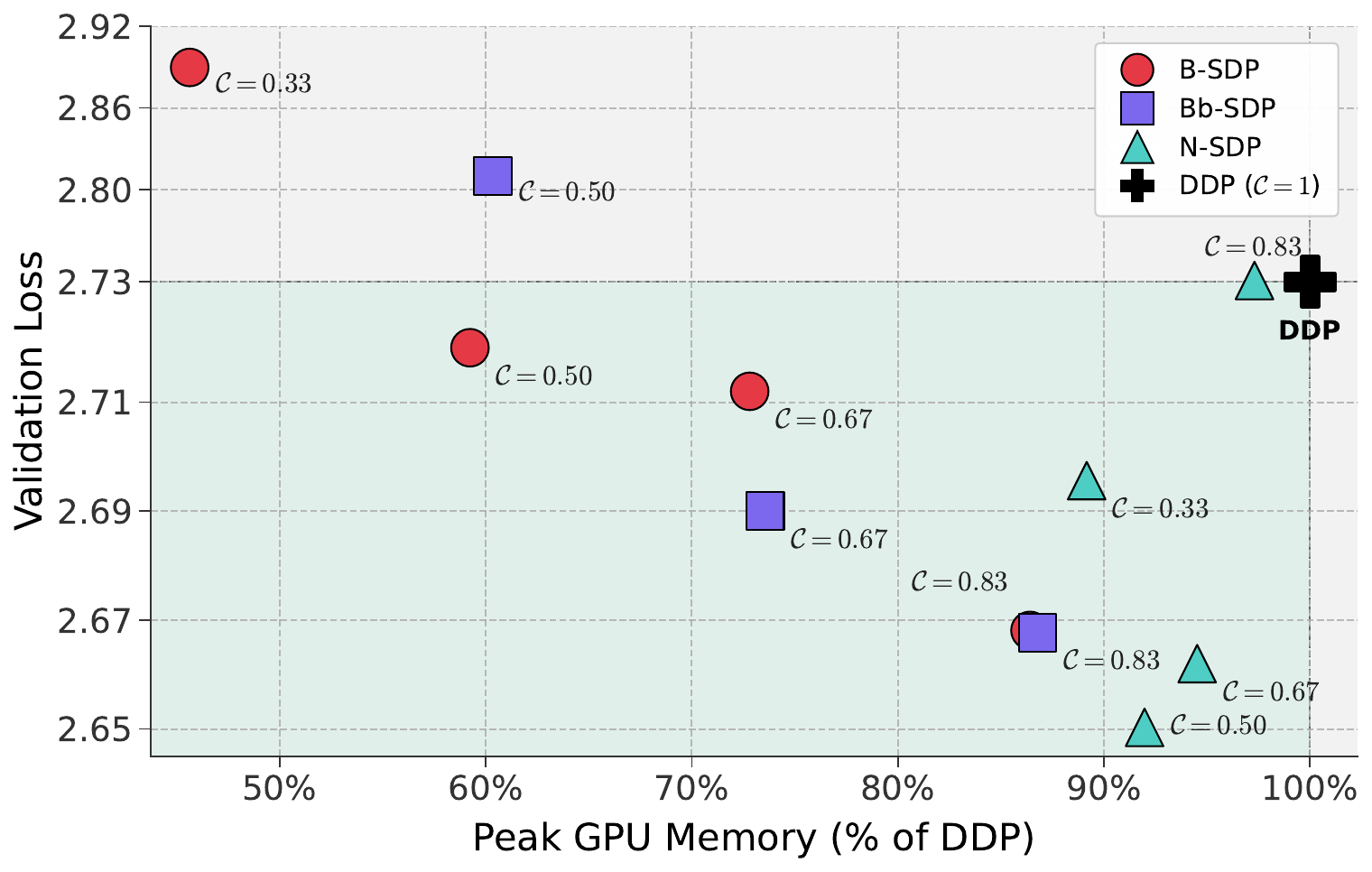}
    \caption{\small 500M LLaMA}
    \label{fig:pareto_500M}
  \end{subfigure}\hfill
  \begin{subfigure}[t]{0.49\linewidth}
    \centering
    \includegraphics[width=\linewidth]{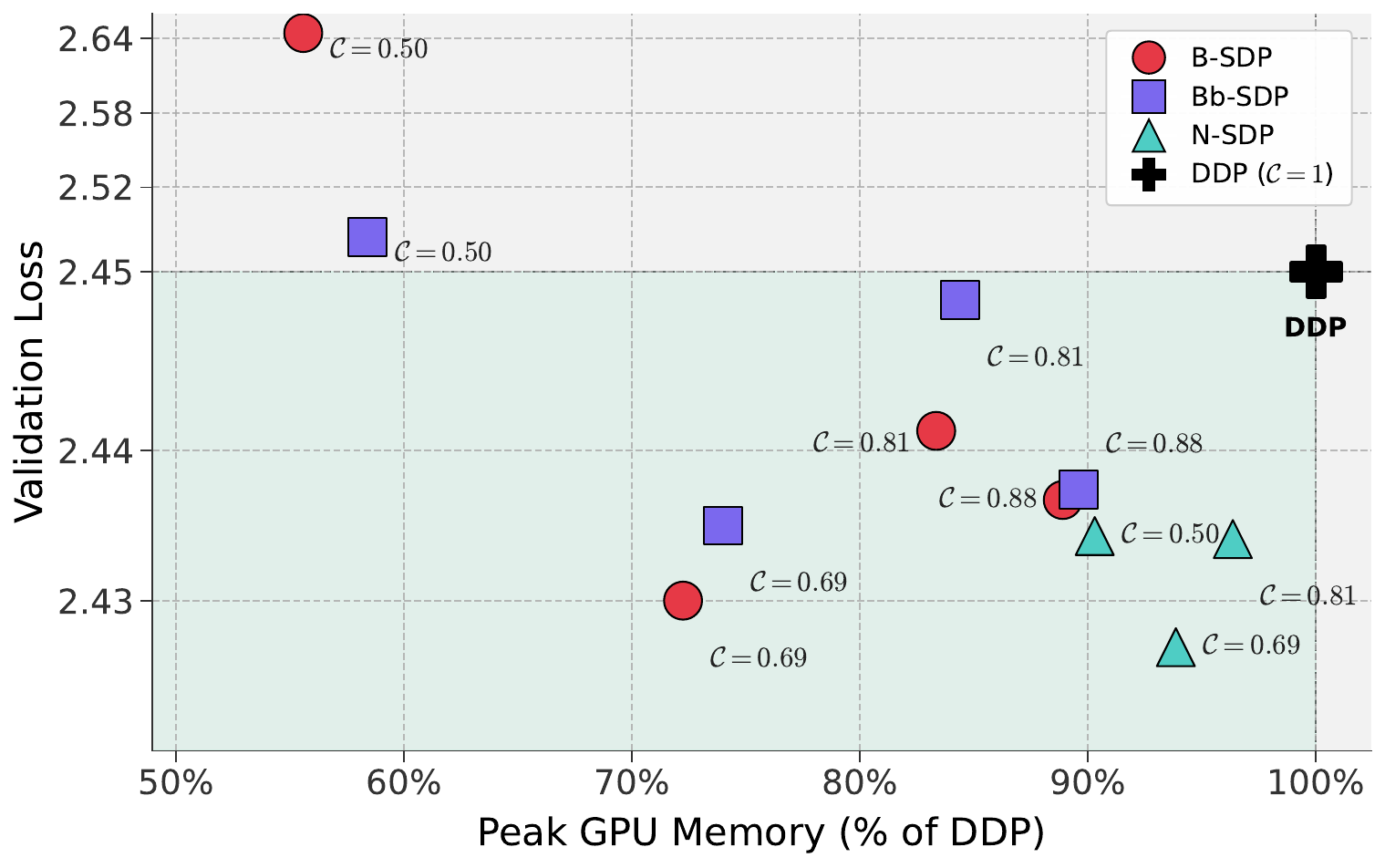}
    \caption{\small 1B LLaMA}
    \label{fig:pareto_1B}
  \end{subfigure}
  \caption{\small \textbf{SDP achieves significant memory reduction at both scales.}
  FLOP-matched LLaMA runs. The coverage ratio \cor\ is the fraction of active transformer components (layers/attention heads) per worker; DDP has \cor$=1$. The shaded \emph{Pareto region} contains configurations that strictly dominate DDP in both peak memory and validation loss (lower the better). The best SDP points reduce maximum peak GPU memory by \hlb{\textbf{40\%}} on 500M (\bsdp, \cor$=0.50$) and \hlb{\textbf{28\%}} on 1B (\bsdp, \cor$=0.69$), while also achieving equal or lower validation loss.}
  \label{fig:pareto_hero}
  \vspace{-15pt}
\end{figure}

In this work, we propose \textbf{\textit{Subnetwork Data Parallelism} (SDP)}, a complementary strategy to model parallelism that reduces per-node memory by distributing the training of model sub-components across nodes. Unlike pipelining, which splits computation into sequential stages, SDP assigns each worker a \textit{\textbf{subnetwork}}, a structurally complete portion of the model (e.g., removing rows and columns of a linear operator) that preserves a full path from input to loss, enabling independent gradient computation without exchanging activations. Each worker optimizes its subnetwork (fixed throughout training) and synchronizes overlapping parameters through stepwise averaging.

We study two instantiations: (i) \emph{\textbf{forward-masked subnetworks}}, which remove both forward and backward computation for a subnetwork, reducing parameters, activations, and gradients for substantial memory savings; and (ii) \emph{\textbf{backward-masked subnetworks}}, where the forward pass uses the full model while masking is applied only in backpropagation, saving gradients and accumulators. The latter retains unbiased gradients and offers a theoretically grounded baseline, while the former provides a practical simplification that empirically improves stability and efficiency. Rather than replicating or fully sharding the model, SDP distributes subnetworks across nodes so each device holds only a fraction of parameters (or gradients/accumulators for backward-masked). Subnetworks are trained independently and synchronized via parameter averaging, yielding a unified model. This significantly lowers memory usage while remaining compatible with intra-node data parallelism and existing systems-level model-parallel techniques.

Unlike pipelining, sharding, or tensor parallelism, our approach modifies the forward and backward computation. Its design rests on three observations. First, overlapping parameter assignments with periodic averaging maintain partial synchronization across workers; in forward masked subnetworks, each worker can be viewed as a replica constrained to remain similar through shared overlaps, akin to ensemble alignment strategies~\citep{jolicoeur2023population, fournier2024wash}. Second, in the backward masked regime, the forward pass uses the full model while sparsity is applied only during backpropagation. In this case, gradient estimates remain unbiased, and deviations from full DP are governed by mask connectivity, providing a principled baseline with theoretical guarantees. Third, subnetworks reduce per-iteration time thus decreased per-iteration convergence rates (which we demonstrate theoretically) can be offset by increasing iterations in a FLOP-matched manner. Our primary contributions in this work are:
\begin{itemize}
    \item We propose a novel distributed training paradigm: \textbf{\textit{Subnetwork Data Parallelism} (SDP)} enabling memory efficient distributed training. We instantiate SDP with two subnetwork-construction strategies (block-level layer removal and neuron/channel removal) and provide a theoretical basis linking the convergence of backward-masked SDP to a spectral-gap condition on the mask graph.
    \item On 1B-parameter LLaMA pre-training (FLOP-matched on FineWeb), \textbf{SDP substantially reduces peak GPU memory and per-step communication while preserving validation loss}, and it composes with FSDP ZeRO-3 and activation checkpointing to push both below either baseline alone.
    \item \textbf{SDP transfers and generalizes cleanly to image classification} (Swin Transformer and ResNet-18 / WideResNet-18 on CIFAR-10/100), where it matches or exceeds DDP accuracy at a fraction of DDP's memory budget, demonstrating that SDP is a drop-in data-parallel alternative whose advantage grows as memory and bandwidth tighten.
\end{itemize}
\begin{figure*}[t]
 \centering
{\includegraphics[trim=5 5 5 5, clip, width=1\textwidth]{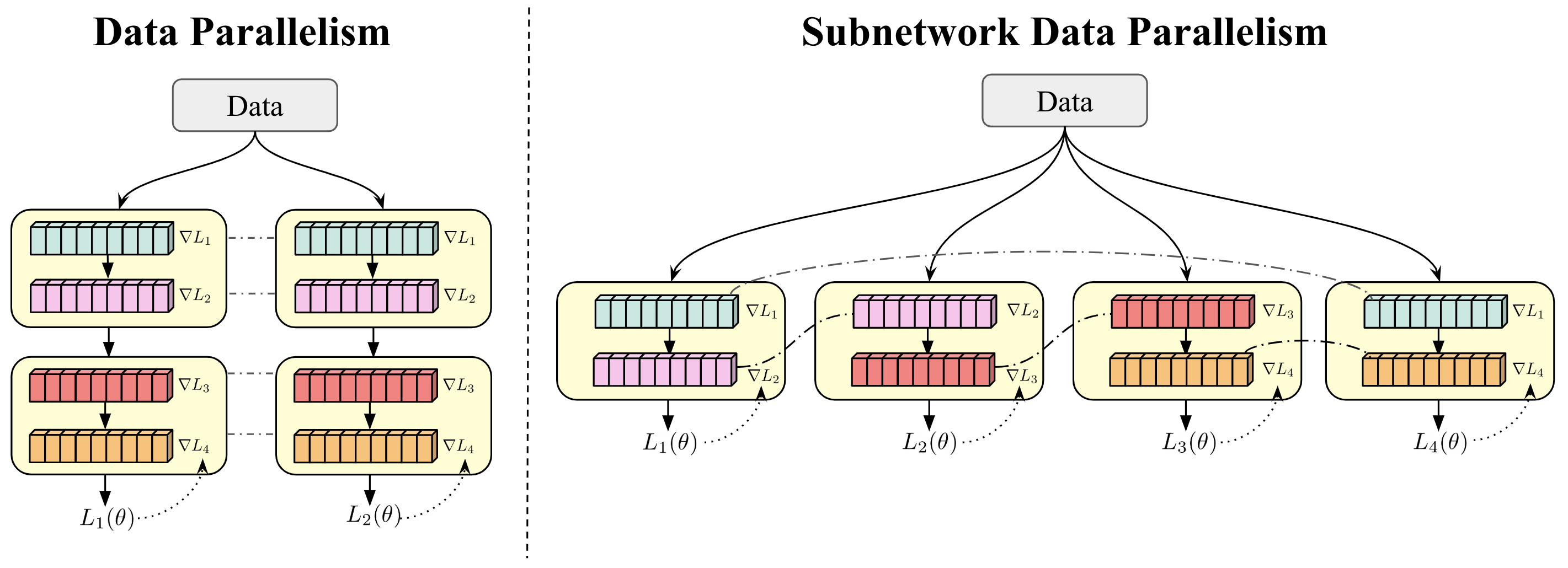}}
\caption{\small \textbf{Data Parallelism (DDP) vs.\ Subnetwork Data Parallelism (SDP).}
\emph{Left:} In data parallelism each GPU hosts a full replica, computes all layer gradients
$\{\nabla L_1,\nabla L_2,\nabla L_3,\nabla L_4\}$, and all-reduces \emph{all} parameters each step; per-GPU memory is approximately the full model (parameters $+$ gradients $+$ optimizer state $+$ activations).
\emph{Right:} In SDP each GPU trains an end-to-end \emph{subnetwork} (a subset of layers/neurons) with a local loss $L_k(\theta)$; only gradients of \emph{shared} parameters are synchronized via masked averaging (dashed arcs). For a \textbf{parameter density per GPU} \textbf{\cor} $\in$ (0,1], both memory and communication per GPU scale as $\approx$ \cor$\times$  DP, with no cross-GPU activation exchange. This enables fitting larger models or longer sequences under the same hardware budget and improves scalability when bandwidth or memory are bottlenecks; when \cor$=1$ (all parameters on all GPUs), SDP reduces to standard DDP.
}
\label{fig:subnet}
\vspace{-15pt}
\end{figure*}

\section{Related Work}

\textbf{Pipeline parallelism:}
Pipeline parallelism reduces memory bottlenecks by splitting the model across devices. \citet{huang2019gpipe, petra} partition layers and pipeline micro-batches, while Mesh-TensorFlow~\citep{shazeer2018mesh} and Megatron-LM~\citep{shoeybi2019megatron} shard weights and activations within layers. These methods overcome memory limits but require high bandwidth interconnects and still suffer from pipeline bubbles and load imbalance. Another line of work explores parallel layer training via auxiliary local losses~\citep{belilovsky2020decoupled}.

\textbf{Fully Sharded and Zero Redundancy Approaches:}
To reduce memory inefficiencies in data parallelism, methods like Fully Sharded Data Parallel (FSDP)~\citep{fsdp} and ZeRO~\citep{rajbhandari2020zero} partition parameters, gradients, and optimizer states across devices. These approaches, supported by frameworks such as DeepSpeed~\citep{rasley2020deepspeed}, greatly lower per-device memory but still incur substantial communication, especially during gradient synchronization, leading to higher overhead and latency.
\paragraph{Ensemble and SWARM Learning:} Recent work~\citep{fournier2024wash,jolicoeur2023population} shows the benefits of training multiple related models in parallel. Forward Subnetwork Masking can be seen as a similar framework, training diverse yet connected models while enforcing alignment, but with the added advantage of reduced per-iteration compute and memory. SWARM~\citep{ryabinin2023swarm} addresses model parallelism limits by assigning multiple devices to each pipeline stage and routing samples efficiently. In contrast, it still requires activation communication over potentially low-bandwidth links, whereas our subnetwork approach reduces all communication to parameters or gradients while maintaining only data parallelism across nodes.
\paragraph{Federated Learning and Dropout-Based Subnetwork Training:}
Federated learning frameworks~\citep{konevcny2016federated} train models across decentralized data sources, often addressing the non-IID challenge. Several works explore training subnetworks per device in this setting~\citep{caldas2018expanding,horvath2021fjord,guliani2022enabling,wen2022federated,alam2022fedrolex}, but with different goals and methodology. These focus on reducing communication and device compute, whereas our aim is to lower memory requirements, critical for training large models on memory limited GPUs. Communication load, by contrast, is well studied and can be mitigated through multi-step training and compression methods~\citep{reddi2021adaptive,Diloco,wang2023cocktailsgd}.

FedRolex \citep{alam2022fedrolex} and HeteroFL \citep{diao2020heterofl} vary model size across clients to address device heterogeneity in compute and memory, assigning subnetworks via channel-level dropout in CNNs. Our work instead targets a homogeneous setting, aiming to lower per-node memory through a more general subnetwork assignment strategy. Moreover, while these methods operate under privacy and heterogeneity constraints, often leaving each client with only small datasets, we assume each worker can access the full dataset. This avoids issues of heterogeneity or overfitting. Finally we consider block level masks and LLMs.

In these works, assigned masks are dynamic which adds significant communication and coordination overhead in a non-federated setting where wall-clock time is critical. \citet{yuan2019distributed} studied dynamic non-overlapping subnetworks with local SGD, whereas our fixed masks simplify the system and enable efficient forward and backward strategies. The overlapping nature of fixed subnetworks is key: shared assignments keep parameters aligned through averaging, and our analysis shows that convergence quality degrades with reduced overlap. Moreover, while \citet{yuan2019distributed} focused on MLPs, our method scales to standard architectures for image classification and large-scale language pre-training. To the best of our knowledge, none of these work considers masking only the backward pass. \citet{fagnou2025accelerated} examined skipping backward blocks in residual networks to speed up training, but did not address distributed settings or memory reduction in the FLOP-matched regime.

\section{Method}
\label{sec:method}
We introduce a distributed training framework that enhances memory efficiency in gradients, activation, and weight storage by defining a communication pattern between workers and model parameters. First we describe a generic multi-worker masking framework which considers fixed masks on parameters, gradients or both in the forward and backward pass of training. Then we specialize this to structured masks that yield benefits in memory and per-iteration speed.

\subsection{Forward and Backward Masking}
\label{meth:fwd_bwd_mask}
\paragraph{Gated coordinates.}
Consider a distributed setting with $n$ workers (GPUs). Let $J$ be an index set of \emph{coordinates} of the model; we use “coordinate” to refer to an index of the parameter vector $\boldsymbol{\theta}$ and, by the same index set, the corresponding coordinate of its gradient $\nabla_{\boldsymbol{\theta}}\mathcal{L}$. Each coordinate $j \in J$ is assigned to a subset of workers, with overlaps allowed. This assignment is encoded by a binary masking matrix $\mathbf{m} \in \{0,1\}^{n \times |J|}$, where $m_{i,j}=1$ means worker $i$ is responsible for coordinate $j$. Using this mask, we define the \emph{gated parameters} by elementwise multiplication with the global parameter vector:
\begin{equation}
\forall i \leq n, \forall j \in J,\quad (\mathbf{m} \odot \boldsymbol{\theta})_{i,j} \triangleq m_{i,j}\,\theta_j. \tag{gate}
\end{equation}
Appendix \ref{mak-construction} describes a procedure that allows to construct a matrix that often satisfies in practice $\sum_{j=1}^m m_{i,j}=p$ and $\sum_{i=1}^n m_{i,j}=c$.
Given per--worker gradients $\mathbf{g}_1, \ldots, \mathbf{g}_n \in \mathbb{R}^{|J|}$, we define the \emph{gated average} for $j \in J$ as the column-wise average over assigned workers:
\begin{equation}
\bar{\mathbf{m}}(\mathbf{g}_1,\ldots,\mathbf{g}_n)_j \triangleq
\frac{1}{c} \sum_{i=1}^n \mathbf{m}_{i,j} (\mathbf{g}_i)_j
= \frac{1}{c} \sum_{i=1}^n m_{i,j} (\mathbf{g}_i)_j. \tag{average}
\end{equation}
\paragraph{Forward and Backward Masking} We assume that we have access to two masks $\mathbf{m}_{\mathrm{fwd}}, \mathbf{m}_{\mathrm{bwd}}$. At step $t$, worker $i$ draws a mini-batch $\mathcal{B}_i^{(t)}$ using a forward mask $\mathbf{m}_{\mathrm{fwd}}$. The forward pass evaluates the loss at
\[
\boldsymbol{\theta}_i^{(t)} = (\mathbf{m}_{\mathrm{fwd}} \odot \boldsymbol{\theta}^{(t)})_i,
\qquad
\mathbf{g}_i^{(t)} = \nabla_{\boldsymbol{\theta}}\mathcal{L}\!\left(\boldsymbol{\theta}_i^{(t)};\,\mathcal{B}_i^{(t)}\right).
\]
Note at this stage, that if $(\mathbf{m}_{\mathrm{fwd}})_{i,j}=0$ then $(\mathbf{g}_i^{(t)})_j=0$. The backward pass applies the backward aggregation mask on the resulting gradients $\mathbf{m}_{\mathrm{bwd}}$ component wise:
\[
\hat{\mathbf{g}}^{(t)} = \bar{\mathbf{m}}_{\mathrm{bwd}}\bigl({\mathbf{g}}_1^{(t)},\ldots,\mathbf{g}_n^{(t)}\bigr),
\]
followed by the optimizer update: 
$
\boldsymbol{\theta}^{(t+1)} = \boldsymbol{\theta}^{(t)} - \text{OptUpdate}\bigl(\hat{\mathbf{g}}^{(t)}, \mathbf{s}\bigr).
$

We consider two variants of this under the Subnetwork DP framework
\begin{itemize}
  \item \textit{Forward-masking:} $\mathbf{m}_{\mathrm{fwd}}=\mathbf{m}$ and $\mathbf{m}_{\mathrm{bwd}}=\mathbf{m}$. The model is evaluated at masked parameters; activations are gated and memory-saving, but gradients reflect this masked forward.
  \item \textit{Backward-masking:} $\mathbf{m}_{\mathrm{fwd}}=\mathbf{m}^{\mathrm{uni}}$ and $\mathbf{m}_{\mathrm{bwd}}=\mathbf{m}$. The model is evaluated at full $\boldsymbol{\theta}^{(t)}$ (no forward bias); sparsity appears only in backprop/aggregation.
\end{itemize}
Choosing $\mathbf{m}_{\mathrm{fwd}}=\mathbf{m}^{\mathrm{uni}}$ keeps activations identical across workers (the standard mini-batch setting).
\paragraph{Convergence in the $L$-smooth case (Backward-masking, simple).}
Assume $f:\mathbb{R}^{|J|}\!\to\!\mathbb{R}$ is $L$-smooth. In the backward-masked (BM) setting we take
\[
\mathbf{m}_{\mathrm{fwd}}=\mathbf{m}^{\mathrm{uni}},\quad \mathbf{m}_{\mathrm{bwd}}=\mathbf{m},
\]
so the forward pass is unmasked and only the backward/aggregation is masked.
Each worker $i$ computes a stochastic gradient $\mathbf{g}_i^{(t)}$ at $\boldsymbol{\theta}^{(t)}$ with
\begin{equation}
\textstyle
\begin{aligned}
\mathbb{E}\bigl[\mathbf{g}_i^{(t)} \mid \boldsymbol{\theta}^{(t)}\bigr] =\nabla f(\boldsymbol{\theta}^{(t)}),\\ \quad
\mathbb{E}\bigl[
\|\mathbf{g}_i^{(t)} - \nabla f(\boldsymbol{\theta}^{(t)})\|^2\mid \boldsymbol{\theta}^{(t)}
\bigr] \le \sigma^2 ,
\end{aligned}
\end{equation}
and we aggregate by masked averaging
\[
\widehat{\mathbf{g}}^{(t)} \;=\; \bar{\mathbf{m}}\!\left(\mathbf{g}_1^{(t)},\ldots,\mathbf{g}_n^{(t)}\right).
\]
Let $\mathbf{g}_{\mathrm{uni}}^{(t)} \!=\! \bar{\mathbf{m}}^{\mathrm{uni}}(\mathbf{g}_1^{(t)},\ldots,\mathbf{g}_n^{(t)})$ and define the masking error $\boldsymbol{\delta}^{(t)}\!\triangleq\!\widehat{\mathbf{g}}^{(t)}-\mathbf{g}_{\mathrm{uni}}^{(t)}$. By linearity of expectation, both $\widehat{\mathbf{g}}^{(t)}$ and $\mathbf{g}_{\mathrm{uni}}^{(t)}$ are unbiased for $\nabla f(\boldsymbol{\theta}^{(t)})$ under BM. The update is
\[
\boldsymbol{\theta}^{(t+1)}=\boldsymbol{\theta}^{(t)}-\eta\,\widehat{\mathbf{g}}^{(t)}.
\]
In fact, if we define $\rho=\Vert \frac 1n\mathbf{m}^{\mathrm{uni}}-\frac 1c\mathbf{m}\Vert_{\mathrm{Frob}}$,
then we have the following lemma that characterizes the convergence speed in the \(L\)-smoothness setting:

\begin{theorem}[SGD rate under Backward-masking]\label{thm:bm-simple}
If $f$ is $L$-smooth and $\eta \le \frac{1}{2L(1+n\rho^2)}$, then for any $T\ge1$,
\begin{equation}
\textstyle
\frac{1}{T}\sum_{t=0}^{T-1}\mathbb{E}\|\nabla f(\boldsymbol{\theta}^t)\|^2
\le
\frac{2(f(\boldsymbol{\theta}^{0})-f^\star)}{\eta T}
+ L\eta\,\sigma^2\rho^2,
\end{equation}

where $f^\star \triangleq \inf_{\boldsymbol{\theta}} f(\boldsymbol{\theta})$ and $\rho^2=\Vert \frac 1n\mathbf{m}^{\mathrm{uni}}-\frac 1c\mathbf{m}\Vert^2_{\mathrm{Frob}}$ is the distance to the uniform masking.
\end{theorem} We defer the proof in \autoref{appn:proof}. 
\if false
\paragraph{Remarks on the forward-masked variant.}
If $\mathbf{m}_{\mathrm{fwd}}=\mathbf{m}$ (FM), gradients are evaluated at masked parameters $\boldsymbol{\theta}_i^{(t)}=(\mathbf{m}\odot \boldsymbol{\theta}^{(t)})_i$. By $L$-smoothness of $f_i$,
\[
\bigl\|\nabla f_i(\boldsymbol{\theta}_i^{(t)}) - \nabla f_i(\boldsymbol{\theta}^{(t)})\bigr\|
\;\le\; L\,\bigl\|(\mathbf{1}-\mathbf{m}_{i,\cdot})\odot \boldsymbol{\theta}^{(t)}\bigr\|.
\]
Thus forward masking introduces an additional forward-bias term proportional to how much each worker zeros out, which propagates into the same descent analysis as an extra additive error on top of the aggregation deviation controlled by $\rho$. In practice we find that the the solution reached by forward masking can be competitive and sometimes exceed the principled backward masking while having a lower overhead.
\fi
\subsection{Subnetwork Data Parallelism with Structured Mask Construction}
\label{meth:sdp_constr}

Our framework instantiates \textbf{\textit{Subnetwork Data Parallelism (SDP)}} by employing \emph{structured masks}, which remove entire parameter groups including parameters, gradients, accumulators, and activations from each worker. This yields substantial memory savings and per-iteration speedups, offsetting slower convergence while preserving the efficiency benefits of subnetworks. We introduce two strategies for instantiating \textit{subnetworks}: \textbf{\textit{Neuron-Level SDP (N-SDP)}}, based on dropout~\citep{srivastava2014dropout} for fully connected and convolutional layers, and \textbf{\textit{Block-Level SDP (B-SDP)}}, inspired by stochastic depth~\citep{huang2016deep} for residual architectures.

\paragraph{Neuron-Level SDP (N-SDP).}
Through \nsdp\ we instantiate subnetworks by selectively removing neurons in fully connected layers (or channels in convolutional layers). For two successive layers $(W^l, W^{l+1})$ with $W^l: \mathbb{R}^{d_{l-1}} \to \mathbb{R}^{d_l}$, dropping outputs of layer $l$ naturally removes the corresponding inputs of layer $l{+}1$. For simplicity, we restrict to \emph{forward masking}, where the same mask is applied in both directions ($\mathbf{m}_{\mathrm{fwd}} = \mathbf{m}_{\mathrm{bwd}} = \mathbf{m}$; see Section~\ref{meth:fwd_bwd_mask}). Applying $m^l$ to layer $l$ thus induces a consistent $m^{l+1}$ on layer $l{+}1$. As a result,
\[
(\mathbf{m}^l \odot W^l, \, W^{l+1})
\quad \text{and} \quad
(\mathbf{m}^l \odot W^l, \, \mathbf{m}^{l+1} \odot W^{l+1})
\]
produce identical outputs. For example, if $W^l, W^{l+1} \in \mathbb{R}^{d \times d}$ and we mask a subset $J_{\mathrm{mask}} \subset \{1,\ldots,d\}$ of output neurons, then setting
\begin{equation}
\begin{aligned}
m^l_{jk} &= 0,
&\quad j \in J_{\mathrm{mask}},\; k \in \{1,\ldots,d\}, \\
m^{l+1}_{kj} &= 0,
&\quad j \in J_{\mathrm{mask}},\; k \in \{1,\ldots,d\}.
\end{aligned}
\end{equation}
ensures that both layers remain consistent under the masking operation.
\paragraph{Block-Level SDP (B-SDP).} Here, subnetworks are formed by removing entire blocks, particularly in architectures with skip connections. Let the model have $L$ blocks $\{B^{1},\ldots,B^{L}\}$ with parameters $\boldsymbol{\theta}^{(l)}$. Each block has a binary mask $m^{(l)} \in \{0,1\}$ denoting whether it is active. When $m^{(l)}=0$, the block is skipped and its parameters excluded. In residual architectures (e.g., ResNets), this reduces to the identity mapping via the skip path, ensuring valid representations even when blocks are dropped.
Formally, a residual connection and its respective masked computation at block $l$ is as follows:  
\begin{equation}
B^{(l)}(\mathbf{x}) + \mathbf{x}, \quad  \text{and } \quad
    \hat{B}^{(l)}(\mathbf{x})
    = m^{(l)} B^{(l)}(\mathbf{x}) + \mathbf{x}.
\end{equation}
We also consider the more general case of \emph{backward masking}, where $\mathbf{m}_{\mathrm{fwd}}=\mathbf{m}^{\mathrm{uni}}$ and $\mathbf{m}_{\mathrm{bwd}}=\mathbf{m}$ as explained in Section~\ref{meth:fwd_bwd_mask}. We refer to this instantiation as \bbsdp\, where the block may be active during the forward pass but omitted during the backward pass.

\paragraph{Memory, compute, and communication cost}
Let $N$ be the total parameter count and \cor $\in (0,1]$ the per-worker density (fraction of coordinates selected by the mask). With fp32 parameters (4 bytes), fp32 gradients (4 bytes), and Adam accumulators in fp32 (8 bytes), standard DP requires $\approx 16N$ bytes per-worker. This is the storage configuration under mixed precision training through \texttt{torch.amp.autocast(bf16)}, which casts inputs to bf16 for compute only and leaves storage dtypes unchanged. In \emph{Forward masking:} only the \cor$N$ coordinates materialize parameters, gradients, and accumulators, using $\approx 16$\cor$N$ bytes; activations and compute also scale $\approx$ \cor\ for structured masks (channel/block level). In \emph{Backward masking:} the full forward is computed, but gradients and accumulators are stored only for the \cor$N$ active coordinates, giving $(4+12$\cor$)N$ bytes. Activations scale $\approx$ \cor.  Block forward masking is illustrated in Figure~\ref{fig:subnet} and compared with DDP pipelining.

\paragraph{Communication cost}
Communication cost is also reduced under SDP. In ring all-reduce, each worker with $N$ parameters sends and receives about $2N$ scalars per step, whereas SDP synchronizes only the \cor $N$ active coordinates, reducing the cost to $\approx 2$\cor $N$. When masks differ across workers, each parameter block is reduced only within its subset of workers; this holds for both forward and backward masking. Gradient compression schemes are well studied in data-parallel settings~\citep{shi2019understanding,xu2021grace}, offering additional savings, but efficient activation compression (e.g., in pipelining or tensor parallelism) remains poorly understood. Thus, SDP can sometimes operate where bandwidth limits preclude other model-parallel methods, while standard techniques (tensor, sharding, pipelining, context) can still be applied within each SDP replica to further reduce memory for large models.

\section{Experiments}
\label{exp}
We now describe our experimental setup for our proposed \textbf{\textit{Subnetwork Data Parallelism} (SDP)} framework on LLM pre-training on FineWeb \citep{penedo2024fineweb} and on image classification on CIFAR-10 / CIFAR-100 \citep{krizhevsky2009learning}. We shall release code for reproducibility at the time of publication.

We define parameter density per worker \cor\ as the ratio of active components ($p$) out of total components ($m$) on each of the $n$ workers. Components can be either computational blocks (for example Basic Block in ResNets and Attention+MLP Block in Transformers) or any parameter vector \(\theta\). This ratio \cor $\,=p/m$ quantifies the sparsity with which a subnetwork is trained across each of the \(n\) workers. For example, \cor\ = \( 6/8\) would imply 6 active components out of total 8 components on each worker (GPUs). By contrast, \(p/m= 8/8=1\) corresponds to the standard distributed-data-parallel (DDP) setup, where each worker trains the full model. We create structured masks via a greedy algorithm as explained in Algorithm \autoref{alg:balanced_assignment} in the appendix.

We evaluate three SDP strategies throughout: \nsdp, \bsdp, and \bbsdp, defined in Section \ref{meth:sdp_constr}. \bsdpr\ and \bbsdpr\ apply block-level masks (mask per transformer or basic block); \nsdpr\ applies attention-head level or channel level masks. All configurations are FLOP-matched against DDP, meaning we scale training iterations inversely with the number of active parameters (eg \cor$=4/8$ for ResNet doubles the target schedule). Consider another example for \bbsdpr\ which performs a full forward pass and receives an additional $1.5\times$ iterations at the same \cor\ to absorb the higher backward cost. In all cases we tune hyperparameters and report results over multiple random seeds.

\subsection{SDP with Large Language Models (LLMs)}
\label{exp:llm}

We pre-train LLaMA-style models~\citep{llama3tok} on FineWeb~\citep{penedo2024fineweb} at three scales: 134M, 500M, and 1B parameters. Token budgets follow Chinchilla scaling~\citep{chinchilla} for the DDP baseline ($20\times$ tokens for 500M/1B). LLMs tend
to have significant memory constraints in practice and thus SDP is a highly pertinent direction for reducing the per-node requirements. \bsdpr\ and \bbsdpr\ use masking per transformer block; \nsdpr\ masks attention heads on $W_Q, W_K, W_V, W_O$. Hyperparameters are detailed in \autoref{app:llm_hparams}. 

\autoref{fig:pareto_hero} shows the 500M and 1B results: at both scales, SDP configurations with more than 50\% active components (\cor$\geq0.50$) strictly dominate DDP in both peak memory and validation loss. At 500M, with just $50\%$ of layers, \textbf{\bsdpr\ saves 40\% per-GPU memory with a lower val.\ loss of 2.72 vs DDP's 2.73.} At 1B scale we see \textbf{28\%} memory savings without sacrificing validation loss. SDP reduces per-device GPU memory because parameters, gradients, optimizer states, and activations for the masked blocks are not materialized on the worker. We also show similar trends for a small 134M model as well in Appendix \autoref{fig:pareto_134M}. Pareto-dominance holds for \bsdpr, \bbsdpr, and \nsdpr, and the dominated region widens with scale. 

To further strengthen our claims, we additionally evaluate the 1B checkpoints on five downstream eval. benchmarks (ARC-Easy, BoolQ, HellaSwag, OpenBookQA, SciQ) using \texttt{lm-evaluation-harness}~\citep{eval-harness}, taking the best validation-loss coverage for each SDP variant. \autoref{tab:downstream_1B} shows that all three variants match DDP within roughly a point on the macro average, with \bsdpr\ giving better average than DDP with its largest per-task gains on OpenBookQA and SciQ; \nsdpr\ posts the lowest validation loss of the four. The main takeaway is the downstream quality preservation: \textbf{SDP recovers DDP-level task performance while delivering the memory and bandwidth savings}. 

\begin{table}[t!]
\centering
\small
\caption{\small \textbf{1B downstream evaluation.} Zero/few-shot accuracy on five benchmarks for DDP and the best-by-average coverage of each SDP variant. All three SDP variants match DDP within $1$pt on the macro average.}
\label{tab:downstream_1B}
\resizebox{\textwidth}{!}{
\begin{tabular}{l c c c c c c c c}
\toprule
\textbf{Method} & \textbf{\cor} & Val. Loss & ARC-E & BoolQ & HellaSwag & OBQA & SciQ & \textbf{Avg} \\
\midrule
DDP                              & $1.00$ & $2.452$ & $\mathbf{0.381}$ & $0.610$ & $0.409$ & $0.344$ & $0.496$ & $0.448$ \\
\bsdp\                           & $0.69$ & $2.437$ & $0.380$ & $\mathbf{0.621}$ & $0.415$ & $\mathbf{0.364}$ & $\mathbf{0.507}$ & $\mathbf{0.457}$ \\
\bbsdp\                          & $0.69$ & $2.437$ & $\mathbf{0.381}$ & $\mathbf{0.621}$ & $0.415$ & $0.320$ & $0.496$ & $0.447$ \\
\nsdp\                           & $0.69$ & $\mathbf{2.434}$ & $0.375$ & $0.600$ & $\mathbf{0.419}$ & $0.344$ & $0.491$ & $0.446$ \\
\bottomrule
\end{tabular}}
\vspace{-15pt}
\end{table}

\subsubsection{Composable Memory Savings}
\label{exp:memory_composition}

\begin{figure}[h]
  \centering
  \begin{subfigure}[t]{0.49\linewidth}
    \centering
    \includegraphics[width=\linewidth]{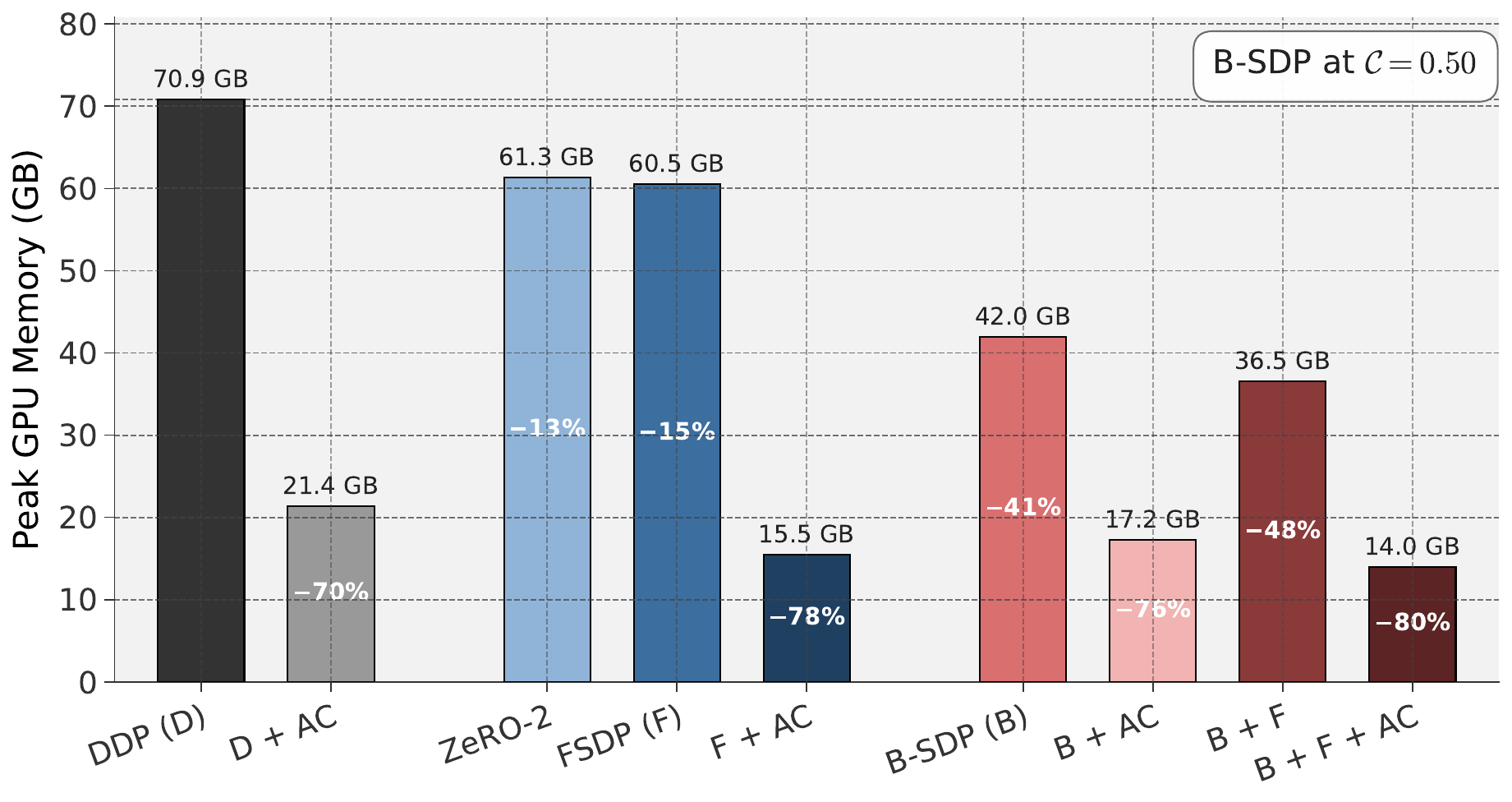}
    \caption{\small 500M LLaMA, batch size=32.}
    \label{fig:mem_500M_bars}
  \end{subfigure}\hfill
  \begin{subfigure}[t]{0.49\linewidth}
    \centering
    \includegraphics[width=\linewidth]{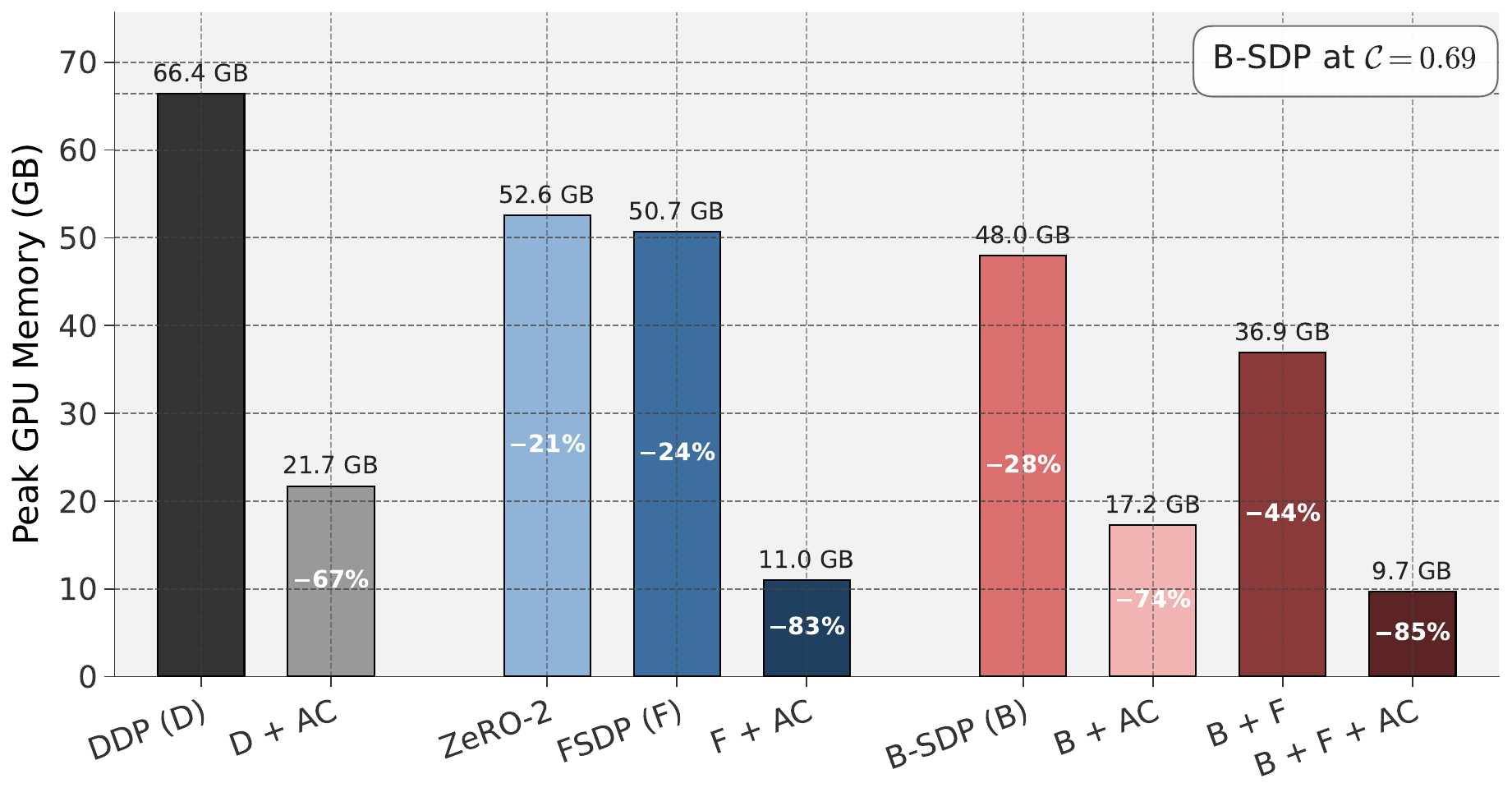}
    \caption{\small 1B LLaMA, batch size=16.}
    \label{fig:mem_1B_bars}
  \end{subfigure}
  \caption{\small \textbf{B-SDP composes with orthogonal memory-saving techniques.} Peak GPU memory per device on 4$\times$H100s, in GB with maximum batch size for each model that could fit DDP. We measure \bsdp\ with \cor$=0.50$ on 500M and \cor$=0.69$ on 1B, the configurations that Pareto-dominate DDP in val. loss (Figure~\ref{fig:pareto_hero}). At 1B scale standalone \bsdp\ saves \hlb{$\boldsymbol{28\%}$} of memory as compared with DDP, and smoothly composes with FSDP and Activation Checkpointing (AC). The three-way composition \textbf{\bsdp+FSDP+AC} reaches $\boldsymbol{9.7}$~\textbf{GB} at 1B (\hlb{$\boldsymbol{-85\%}$}), strictly below FSDP+AC alone ($11.0$~GB at 1B).}
  \label{fig:mem_composition}
  \vspace{-5pt}
\end{figure}
Peak GPU memory in LLM pre-training decomposes into four components namely \textit{parameters, gradients, optimizer state, and activations}. Two standard techniques cut this footprint: Activation Checkpointing (AC), which trades activation storage for recomputation in the backward pass, and FSDP~\citep{fsdp}, which shards the parameter, gradient, and optimizer states across workers and touch only the $16N$ budget. Each of them touches only one of the four components. 

SDP, by contrast, is orthogonal to both. It physically removes a fraction of the transformer blocks or the attention heads from each worker, shrinking all four components in proportion to the coverage ratio (\cor). Now since AC, FSDP and SDP act on disjoint parts of the memory budget, their savings cleanly compose. Figure~\ref{fig:mem_composition} demonstrates this natural memory savings composability at 500M and 1B scale for LLaMA \cite{llama3tok} architecture. At 1B, \bsdpr\ alone cuts peak memory by roughly \textbf{28\%} over DDP, already undercutting FSDP. Combining SDP with FSDP and AC gives maximum memory savings: \textbf{just 9.7~GB per device, 85\% below DDP and lower than FSDP$\,+\,$AC alone (11~GB).}

\subsubsection{Communication Cost}
\label{exp:comm_cost}

We measure end-to-end step time (forward, backward, gradient synchronization, and optimizer) and wall-clock time to the FLOP-matched Chinchilla budget in two regimes: intra-node on $4{\times}$H100 NVLink, and inter-node on $2$ nodes $\times$ $2$ L40 over InfiniBand. Profiling details are in Appendix~\ref{app:comm_profiling}. Table~\ref{tab:comm_1B} reports both regimes side by side. We profile the best-performing SDP configuration from \autoref{fig:pareto_hero}, \bsdpr\ at \cor$=0.69$

Because SDP is FLOP-matched against DDP, per-step compute savings are reinvested as additional iterations. Standalone \bsdpr\ trains for essentially the same wall-clock-time as DDP in both regimes (inter and intra node). The step time however, is meaningfully shorter because each worker now holds fewer parameters (only its assigned subset of transformer blocks), which cuts both the forward and backward compute and the per-step gradient AllReduce volume. For example, \bsdpr\ at \cor$=0.69$ moves $2{,}253$~MB of bus bytes per step against DDP's $3{,}143$~MB, a $28\%$ reduction in communication volume. 

Further, \textbf{B-SDP $+$ FSDP is the fastest-configuration in both regimes.} It delivers a reduction of \textbf{15.4\%} wall-clock time intra-node and \textbf{19.9\%} inter-node relative to DDP, and strictly beats FSDP alone. The gains stem from a compounding reduction in communication volume. Firstly, B-SDP gives each worker only $22$ of the $32$ blocks, the remaining $10$ blocks are absent on each worker and therefore contribute nothing to its forward compute, backward compute, or gradient synchronization. So, the per-step communication traffic scales with the coverage ratio \cor. FSDP then shards the surviving $787$~M across $W{=}4$, leaving $196.8$~M active parameters per GPU and shrinking FSDP's AllGather payload proportionally. This gap widens in inter-node regime where bandwidth pressure amplifies the benefit of cutting per-step bytes.

\begin{table}[t!]
\centering
\small
\caption{\small Communication profiling with step time and end-to-end wall-clock training time for 1B LLaMA at batch size per GPU$=16$, across intra-node (4$\times$H100 NVLink) and inter-node (2 nodes $\times$ 2 L40 over InfiniBand). We profile the best performing SDP configuration, \bsdpr\ at \cor$=0.69$. \textbf{Active (M)} denotes the active parameters per GPU. \textbf{Vol (MB)} is the bus bytes per step. \textbf{Step (ms)} is the full training step including optimizer; \textbf{Walltime (hours)} is the total wall-clock time taken to train the FLOP-matched Chinchilla budget. \textbf{vs.\ DDP} is the wall-clock training time change against the DDP baseline in the same regime (negative = faster). }
\label{tab:comm_1B}
\resizebox{\textwidth}{!}{
\begin{tabular}{l r r r r r r r r}
\toprule
 & & & \multicolumn{3}{c}{\textbf{Intra-node} ($4{\times}$H100, mbs$=16$)} & \multicolumn{3}{c}{\textbf{Multi-node} ($2{\times}2$ L40, mbs$=4$)} \\
\cmidrule(lr){4-6} \cmidrule(lr){7-9}
\textbf{Method} & \textbf{Active (M)} & \textbf{Vol (MB)} & \textbf{Step (ms)} & \textbf{Walltime (h)} & \textbf{vs.\ DDP$_\text{torch}$} & \textbf{Step (ms)} & \textbf{Walltime (h)} & \textbf{vs.\ DDP$_\text{torch}$} \\
\midrule
 DDP   & $1{,}098.7$ & $3{,}143$ & $334.14$ & $14.16$ &      & $383.55$ & $65.03$ &      \\
DDP $+$ ActCkpt                   & $1{,}098.7$ & $3{,}143$ & $433.93$ & $18.39$ & $+29.9\%$ & $454.95$ & $77.13$ & $+18.6\%$ \\
FSDP ZeRO-3                       & $274.7$     & $4{,}568$ & $330.20$ & $14.00$ & $-1.1\%$  & $453.12$ & $76.82$ & $+18.1\%$ \\
FSDP ZeRO-3 $+$ ActCkpt           & $274.7$     & $4{,}568$ & $440.06$ & $18.65$ & $+31.7\%$ & $559.04$ & $94.78$ & $+45.7\%$ \\
\midrule
\bsdp\              & $787.3$     & $2{,}253$ & $241.45$ & $14.28$ & $+0.8\%$  & $276.99 $ & $65.55 $ & $+0.8\% $ \\
\bsdp\ $+$ ActCkpt  & $787.3$     & $2{,}253$ & $302.99$ & $17.92$ & $+26.6\%$ & $328.47 $ & $77.72 $ & $+19.5\% $ \\
\rowcolor{lightblue} \bsdp\ $+$ FSDP Z3  & $196.8$     & $3{,}232$ & $\mathbf{202.49}$ & $\mathbf{11.98}$ & $\mathbf{-15.4\%}$ & $\mathbf{220.58}$ & $\mathbf{52.05}$ & $\mathbf{-19.9\%}$ \\
\bsdp\ $+$ FSDP Z3 $+$ ActCkpt  & $196.8$ & $3{,}232$ & $262.63$ & $15.54$ & $+9.7\%$ & $270.91$ & $63.89$ & $-1.8\%$ \\
\bottomrule
\end{tabular}}
\vspace{-10pt}
\end{table}
\subsection{SDP with Image Classification}
\label{exp:IMC}

We also apply SDP to two image-classification architectures with different inductive biases: the Swin Transformer \cite{liu2021Swin} (Swin-Tiny, $d=12$ transformer blocks) and the ResNet-18 / WideResNet-18 CNN \cite{he2016deep}. The three SDP strategies (Section \ref{exp}) adapt to each architecture as follows. On Swin, \bsdpr\ and \bbsdpr\ mask whole transformer blocks (varying the the active components $p \in \{10, 9, 8, 6, 5\}$ out of $12$ blocks), and \nsdpr\ masks attention heads in stages 2--4 (Heads in 4 stages: $\{3, 6, 12, 24\}$), excluding stage 1 as its low head count limits pruning capacity without harming representation learning. On ResNet, \bsdpr\ and \bbsdpr\ mask whole basic blocks (by varying $p \in \{8, 7, 6, 5, 4, 3\}$ out of $8$ basic blocks), and \nsdpr\ masks channels in the convolutional layers. All experiments train on CIFAR-10 and CIFAR-100 at effective batch size $\mathcal{B}=512$ across $n=8$ workers, FLOP-matched against DDP; hyperparameters are in \autoref{appn:hyper_swin} (Swin) and \autoref{appn:hyper} (ResNet).

\begin{figure*}[t!]
    \centering
    \subfloat[Swin-Tiny on CIFAR-10 \label{fig:inf:sa}]
    {\includegraphics[trim={4 0 0 0},clip, width=0.5\textwidth]{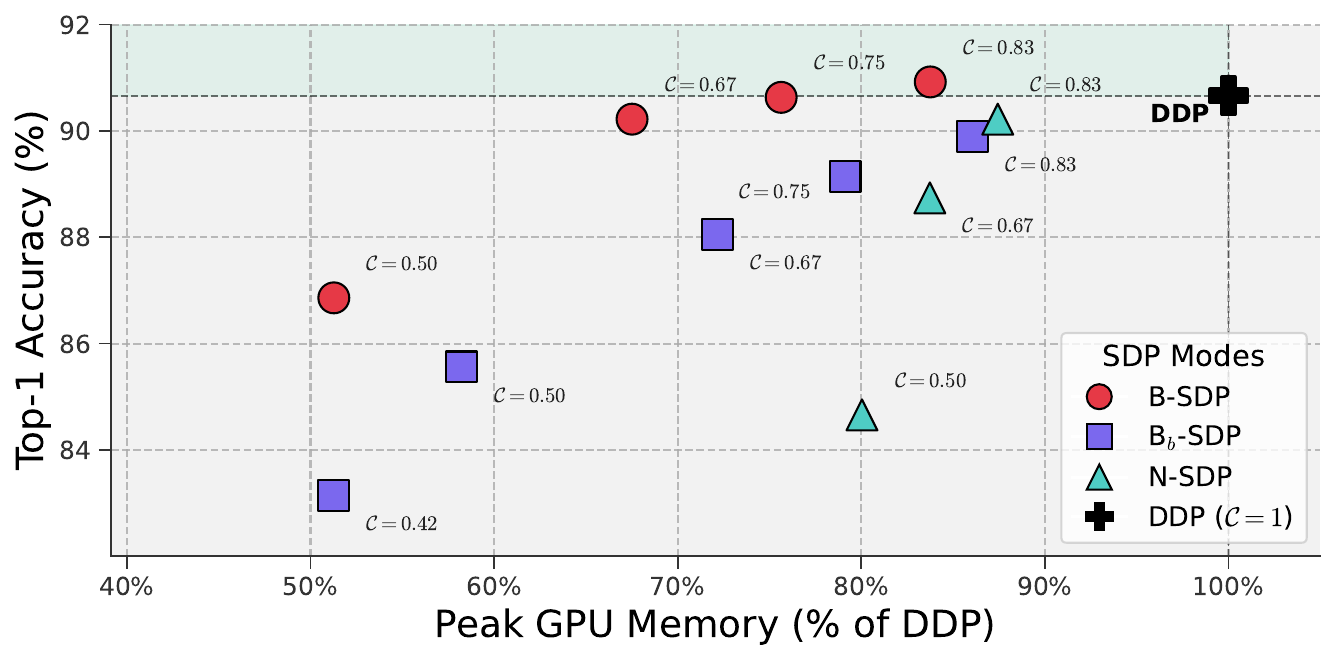}}
    \subfloat[Swin-Tiny on CIFAR-100 \label{fig:inf:sb}]
    {\includegraphics[trim={4 0 0 0},clip, width=0.5\textwidth]{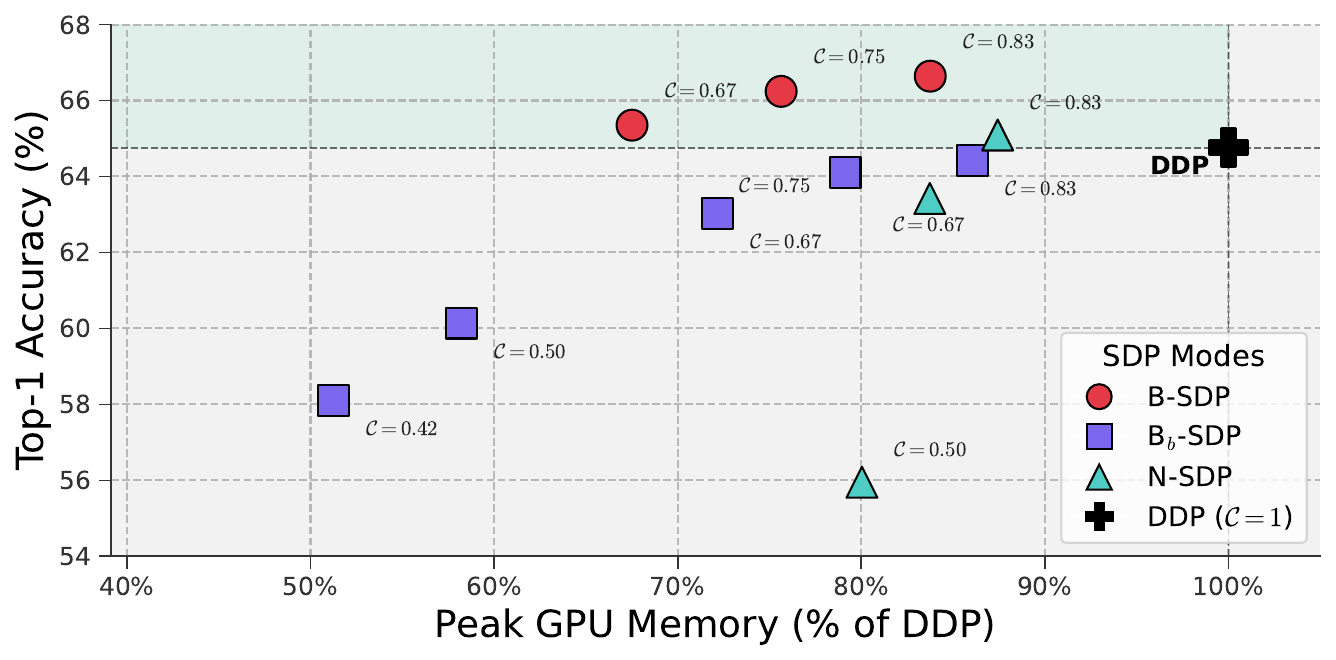}}

    \caption{\small Top-1 accuracy ($\uparrow$) versus fraction of DDP memory for \bsdp, \bbsdp, and \nsdp\ on Swin tiny architecture. All SDP configurations achieve competitive accuracy while using substantially less memory than the DDP baseline, demonstrating a clear efficiency advantage over full-model training. Notably, \bsdp\ consistently delivers the strongest accuracy-memory trade-off (\hlb{\textbf{32\% memory reduction}}), outperforming \nsdp\ and \bbsdp\ at comparable or lower memory fractions.
    }
    \vspace{-5pt}
    \label{fig:swin}
    \end{figure*}

\paragraph{SDP with Swin Transformer architecture:} 
\label{exp:swin}
With just $\boldsymbol{68\%}$ of DDP memory, \textbf{B-SDP matches DDP accuracy} on CIFAR-10 (near $90\%$). On CIFAR-100, it improves by \textbf{2\%} over DDP (from $64.76\%$ to $66.64\%$) at $\boldsymbol{32\%}$ memory reduction as shown in \autoref{fig:swin}. \nsdpr\ drops sharply beyond $80\%$ of DDP memory, suggesting aggressive head masking degrades representational capacity. We also observe that \bbsdpr\ prevents model collapse at high sparsity.

\begin{figure*}[t!]
    \centering
    \subfloat[ResNet-18 on CIFAR-10 \label{fig:inf:a}]
    {\includegraphics[trim={4 0 0 0},clip, width=0.5\textwidth]{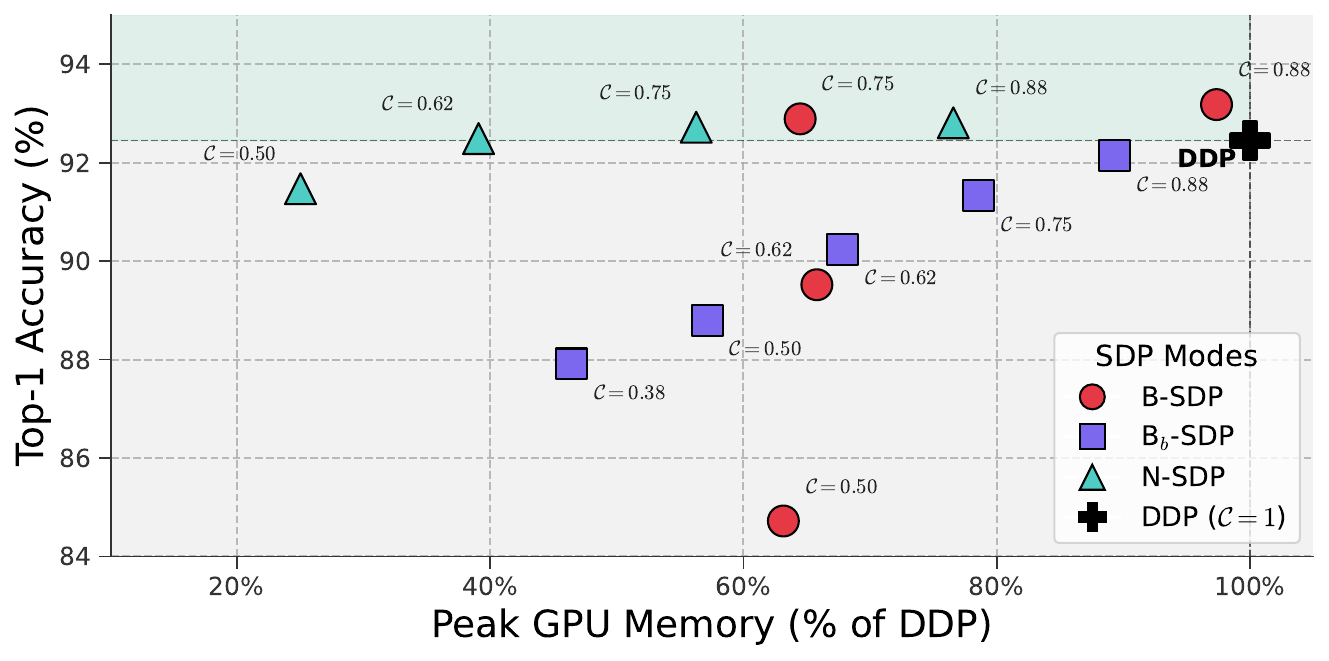}}
    \subfloat[ResNet-18 on CIFAR-100 \label{fig:inf:b}]
    {\includegraphics[trim={4 0 0 0},clip, width=0.5\textwidth]{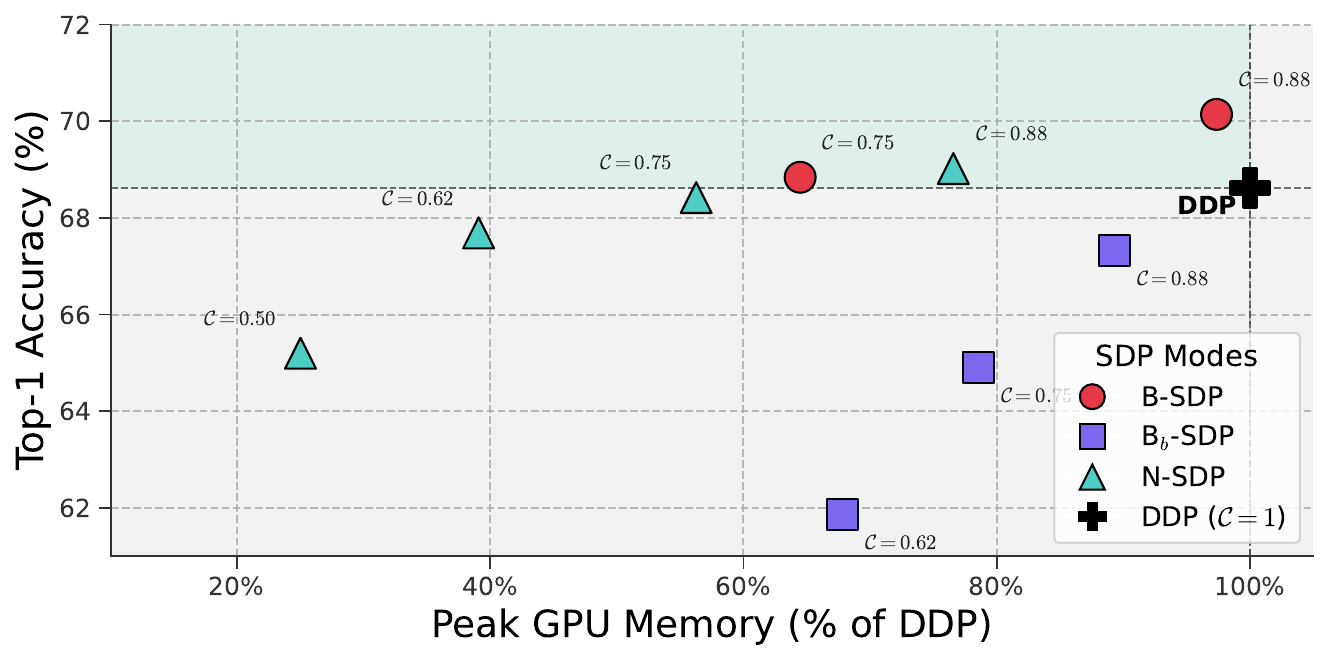}}

    \caption{\small Top-1 accuracy ($\uparrow$) versus fraction of DDP memory for \bsdp, \bbsdp, and \nsdp\ on ResNet-18 across CIFAR-10 and CIFAR-100. SDP configurations consistently achieve competitive or higher accuracy while using substantially less memory (\hlb{\textbf{60\% reduction}}) than the DDP baseline. In particular, \nsdp\ exhibits the strongest accuracy-memory trade-off, outperforming \bsdp\ and \bbsdp\ at comparable or lower memory.
    }
    \label{fig:resnet18}
    \vspace{-15pt}
    \end{figure*}
    
\paragraph{ResNet-18 CNN Architecture:}
\label{exp:resnet}
With SDP \textbf{ResNet-18 retains DDP accuracy using only $\boldsymbol{40\%}$ of DDP memory} on CIFAR-10 (\autoref{fig:resnet18}). \nsdpr\ even \emph{surpasses} DDP on both CIFAR-10 and CIFAR-100 with just $64\%$ memory, while remaining competitive till $56\%$ memory, suggesting a regularization effect from subnetwork training. Across both ResNet variants, \nsdpr\ remains stable with as few as 25\% of DDP memory, while \bsdpr\ and \bbsdpr\ degrades more. WideResNet-18 (WRN-18) and a linear-scheduler robustness check show the same trends (Appendix \autoref{fig:wrn18}, \autoref{tab:rn_wrn_linear}). We attribute the advantage of forward masking at higher \cor\ to two factors: (1) it effectively trains multiple models in parallel, whose diversity improves performance when averaged \cite{fournier2024wash,jolicoeur2023population,streamingdiloco}, and (2) under FLOP matching, forward masking gains more training iterations than backward masking. 

Further, we observe that under severe sparsity (\cor$=3/8$) \bbsdpr\ outperforms both \nsdpr\ and \bsdpr, reaching $87.91\%$ on CIFAR-10 and $58.73\%$ on CIFAR-100 where the other variants collapse. WRN-18 results follow the same trends and are deferred to \autoref{fig:wrn18} in the Appendix. This advantage of backward masking at very low overlap is consistent with the observation in Section 3 that it maintains an unbiased gradient estimate (but at a slower iteration level convergence)


\section{Conclusion}
In this work we present a novel distributed training framework: \textbf{\textit{Subnetwork Data Parallelism} (SDP)} that delivers \textbf{28\%--60\%} memory savings per device while maintaining or even improving performance over DDP. Since SDP physically removes a fraction of model components from each worker, it actually shrinks parameters, gradients, optimizer state, and activations together. Thus providing a natural composability to memory reduction techniques like FSDP, and Activation Checkpointing. This stacking offers a significant reduction of 85\% peak per-device memory at 1B LLaMA pre-training while preserving validation loss and downstream task accuracy. This benefit grows under bandwidth pressure. SDP's per-step byte reduction compounds with FSDP's compute/communication overlap, making the composition the fastest configuration in both intra-node and inter-node setup. Further, SDP generalizes and transfers cleanly to image classification on Swin Transformers and ResNets as well. SDP in this sense, offer a practical third axis of parallelism, complementary to existing techniques and most useful where memory and bandwidth are most constrained.

\section*{Acknowledgements}
This research was supported by NSERC Discovery Grants and FRQNT New Scholar. We acknowledge compute resources provided by the Digital Research Alliance of Canada). EO was supported by PEPR IA on grant
SHARP ANR-23-PEIA-0008 and PEPR NUMPEX on grant  DAIMOS ANR-25-EXNU-0002. Part of this work was
granted access to the JeanZay HPC/AI resources of IDRIS under the allocation AD011015884R1, AD011017481, AD011016641, AD011017661 and AD011017766.

\bibliographystyle{abbrvnat}
\bibliography{main}

\newpage
\appendix
\label{app}

\section{Balanced Mask Construction for SDP}
\label{mak-construction}

Let \(n\) be the number of workers, \(m\) the number of components (can be either a layer or a neuron), and
\(p\in\{0,\dots,m\}\) the per-worker budget.
We construct a binary mask
\[
\mathbf{M}\in\{0,1\}^{n\times m},
\qquad
\mathbf{M}_{i,j}=1 \iff \text{worker } i \text{ is assigned component } j .
\]

Define the column loads \(c_j=\sum_{i=1}^n M_{i,j}\) and the target load \(c^\star=\lceil np/m\rceil\).
The mask is constructed by a (probabilistic) procedure that enforces the exact per-worker budget
\(\sum_{j=1}^m M_{i,j}=p\) for all \(i\),
and heuristically balances column loads by biasing assignments toward under-covered components,
with the goal of approximately minimizing the imbalance
\[
\min \max_{j\in\{1,\dots,m\}} \left|c_j-\frac{np}{m}\right|
\]

Below Algorithm \autoref{alg:balanced_assignment} explains the structured mask construction.

\begin{algorithm}[h]
\caption{Structured Mask Construction}
\label{alg:balanced_assignment}
\begin{algorithmic}[1]

\REQUIRE Number of workers $n$, number of components $m$, active components per worker $p$
\ENSURE Binary matrix $M \in \{0,1\}^{n \times m}$

\STATE Initialize $M \leftarrow \mathbf{0}_{n \times m}$
\STATE Initialize column counts $\mathbf{c} \leftarrow \mathbf{0}_m$
\STATE Set target column count $t \leftarrow \left\lceil \frac{n p}{m} \right\rceil$

\STATE \textbf{Phase 1: Probabilistic balanced assignment}

\FOR{$i = 1$ \textbf{to} $n$}
    \STATE Compute weights $w_j \leftarrow \max(t - c_j, \varepsilon)$ for all $j$
    \STATE Normalize $\{w_j\}$ to probabilities
    \STATE Sample $p$ distinct columns $\mathcal{S}_i$
    \FORALL{$j \in \mathcal{S}_i$}
        \STATE $M_{i,j} \leftarrow 1$
        \STATE $c_j \leftarrow c_j + 1$
    \ENDFOR
\ENDFOR

\STATE \textbf{Phase 2: Column coverage repair}

\FOR{$j = 1$ \textbf{to} $m$}
    \STATE $\Delta \leftarrow t - c_j$
    \IF{$\Delta > 0$}
        \STATE Select $\Delta$ rows $\mathcal{R}'$ s.t. $M_{i,j}=0$ and $\sum_k M_{i,k} < p$
        \FORALL{$i \in \mathcal{R}'$}
            \STATE $M_{i,j} \leftarrow 1$, $c_j \leftarrow c_j + 1$
            \IF{$\sum_k M_{i,k} > p$}
                \STATE Select $k$ with $M_{i,k}=1$ and $c_k > t$
                \STATE $M_{i,k} \leftarrow 0$, $c_k \leftarrow c_k - 1$
            \ENDIF
        \ENDFOR
    \ENDIF
\ENDFOR

\STATE \textbf{return} $M$

\end{algorithmic}
\end{algorithm}

\section{Theoretical Analysis}
\label{appn:proof}
It is a well known result \citep{bottou2010large} that if $f$ is $L$-smooth and $\eta \le \frac{1}{2L}$, then for any $T\ge1$,
\begin{equation}
\textstyle
\frac{1}{T}\sum_{t=0}^{T-1}\mathbb{E}\|\nabla f(\boldsymbol{\theta}^t)\|^2
\le
\frac{2(f(\boldsymbol{\theta}^{0})-f^\star)}{\eta T}
+ L\eta\,\sigma^2\rho^2.
\label{eq:cvg}
\end{equation}
We also require the following variance identity for linear aggregation.

\begin{lemma}[Variance under linear mixing]
\label{lem:var-mixing}
Let $x_1,\dots,x_m \in \mathbb{R}^n$ be i.i.d.\ with $\mathbb{E}[x_i]=x$ and $\operatorname{Cov}(x_i)=\sigma^2 I_n$.
Define the sample mean $\bar x \coloneqq \frac{1}{m}\sum_{i=1}^{m} x_i\in\mathbb R^{n}$ and let
\[
P:(\mathbb R^{n})^{m}\to \mathbb R^{n},\qquad
(Px)_j \coloneqq \sum_{i=1}^{m} P_{ij}\,(x_i)_j
,\quad j=1,\dots,n.
\]
Then
\begin{equation}
\mathbb{E}\,\|Px-\bar x\|_2^2
=\sigma^2\Big\|\,P-\tfrac{1}{m}\mathbf 1_m\mathbf 1_n^\top\Big\|_F^2.
\label{eq:var-mixing}
\end{equation}
\end{lemma}

\paragraph{Proof.}
For each coordinate $j\in\{1,\dots,n\}$,
\[
(Px-\bar x)_j
=
\sum_{i=1}^m\Big(P_{ij}-\tfrac{1}{m}\Big)(x_i)_j.
\]
By independence across $i$ and the assumption $\operatorname{Var}((x_i)_j)=\sigma^2$,
\begin{align*}
\mathbb{E}\,\|Px-\bar x\|_2^2
&=
\sum_{j=1}^n \mathbb{E}\,(Px-\bar x)_j^2
=
\sum_{j=1}^n \operatorname{Var}\!\left(\sum_{i=1}^m\Big(P_{ij}-\tfrac{1}{m}\Big)(x_i)_j\right) \\
&=
\sum_{j=1}^n \sum_{i=1}^m \Big(P_{ij}-\tfrac{1}{m}\Big)^2 \operatorname{Var}((x_i)_j)
=
\sigma^2\sum_{j=1}^n\sum_{i=1}^m\Big(P_{ij}-\tfrac{1}{m}\Big)^2 \\
&=
\sigma^2\Big\|\,P-\tfrac{1}{m}\mathbf 1_m\mathbf 1_n^\top\Big\|_F^2,
\end{align*}
which concludes the proof. \qed

\paragraph{Consequence for \eqref{eq:cvg}.}
By Lemma~\ref{lem:var-mixing}, the effective variance term induced by the linear mixing operator $P$ is
$\sigma^2\big\|\,P-\tfrac{1}{m}\mathbf 1_m\mathbf 1_n^\top\big\|_F^2$.
Therefore, replacing $\sigma^2$ in \eqref{eq:cvg} by this quantity yields the desired bound of Theorem \ref{thm:bm-simple}.

\section{Hyperparameters for ResNet-18 Architecture}
\label{appn:hyper}

All CIFAR-10 and CIFAR-100 experiments in ~\autoref{tab:rn_wrn_combined} and ~\autoref{tab:rn_wrn_linear} are conducted with standard hyperparameters \citep{zhuang2022randomness, cho2025lightweight, petra, wightman2021resnet} an effective batch size of \( \mathcal{B} = 512 \), using 64 samples per GPU across \( n = 8 \) workers. The baseline configuration (\cor $= 1$) is trained for standard 200 epochs. The ResNet experiments employ two learning rate schedules. The first is a cosine annealing schedule with \( \eta_{\text{max}} = 0.2 \) and \( \eta_{\text{min}} = 0.002 \), combined with a linear warm-up over the first 5\% of training iterations to improve convergence stability. The second follows the multi-step linear schedule of~\citet{goyal2017accurate}, where the learning rate is reduced by a factor of 0.1 at predefined milestones. For CIFAR-10, these milestones are at 50\% and 75\% of the total training iterations, while for CIFAR-100 they occur at 30\%, 60\%, and 80\%.

The ResNet experiments use group normalization layers instead of batch normalization layers with 2 groups across all experiments, ensuring that normalization is computed only over active parameters in the subnetwork configurations. Additionally, we adopt a modified Kaiming initialization~\citep{he2015delving}, recalculating the fan-out based on the number of active (unmasked) output units. This adjustment prevents overestimation of activation variance that can occur with standard initialization when masking is applied.

\section{Hyperparameters for Swin-T Architecture}
\label{appn:hyper_swin}
For the experiments on the CIFAR10 and CIFAR100 datasets, we use an effective batch size of \(\mathcal{B} = 512\) across \(n = 8\) workers, training with the AdamW optimizer with momentum for 400 epochs in the baseline DP setting. For configurations with higher sparsity, the training epochs are increased proportionally to ensure FLOP matching, as described in the previous section. As with ResNet-18, we adopt a cosine learning rate schedule with linear warm-up over the first 5\% of iterations, with a peak learning rate of \(\eta_{\text{max}} = 0.0002\) and a minimum learning rate of \(\eta_{\text{min}}\) tending to 1e-7.

\section{Hyperparameters for LLM Architecture}
\label{app:llm_hparams}
We evaluate SDP on LLaMA-style models~\citep{llama3tok} trained on the FineWeb dataset~\citep{penedo2024fineweb} at three scales (134M, 500M, 1B). All three scales share the LLaMA-2 tokenizer ($32\text{K}$ vocabulary), a sequence length of $1024$, and the same optimizer family: AdamW with $\beta_1{=}0.9$, $\beta_2{=}0.95$, weight decay $0.1$, gradient clipping at $1.0$, and a cosine schedule with $10\%$ linear warmup to the peak learning rate down to a floor of $3\mathrm{e}{-6}$. Training runs on $W{=}8$ workers with mixed precision. Per-scale differences are summarized in \autoref{tab:llm_hparams}.
\begin{table}[t!]
\centering
\small
\caption{\small LLM hyperparameters per scale. Token budgets follow Chinchilla scaling~\citep{chinchilla} on the DDP baseline ($20\times$ active parameters for 500M and 1B). For SDP configurations, FLOP-matching is applied per \autoref{exp}. We follow standard cosine learning rate scheduler with warmup.}
\label{tab:llm_hparams}
\begin{tabular}{l c c c}
\toprule
\textbf{Setting} & \textbf{134M} & \textbf{500M} & \textbf{1B} \\
\midrule
Transformer blocks ($L$)            & $12$     & $24$      & $32$      \\
Attention heads                     & $12$     & $24$      & $32$      \\
Hidden size                         & $768$    & $1{,}200$ & $1{,}600$ \\
Total parameters                    & $134$M   & $502$M    & $1{,}098$M \\
Chinchilla token budget (DDP)       & $2.7$B   & $10$B     & $22$B     \\
Sequences length          & $1024$    & $1024$     & $1024$     \\
Global batch size (tokens/step)     & $1.05$M   & $1.05$M   & $1.05$M   \\
Peak learning rate      & $8\mathrm{e}{-3}$ & $4\mathrm{e}{-3}$ & $4\mathrm{e}{-3}$ \\
\bottomrule
\end{tabular}
\end{table}

\begin{figure}[t!]
  \centering
  \includegraphics[width=0.65\linewidth]{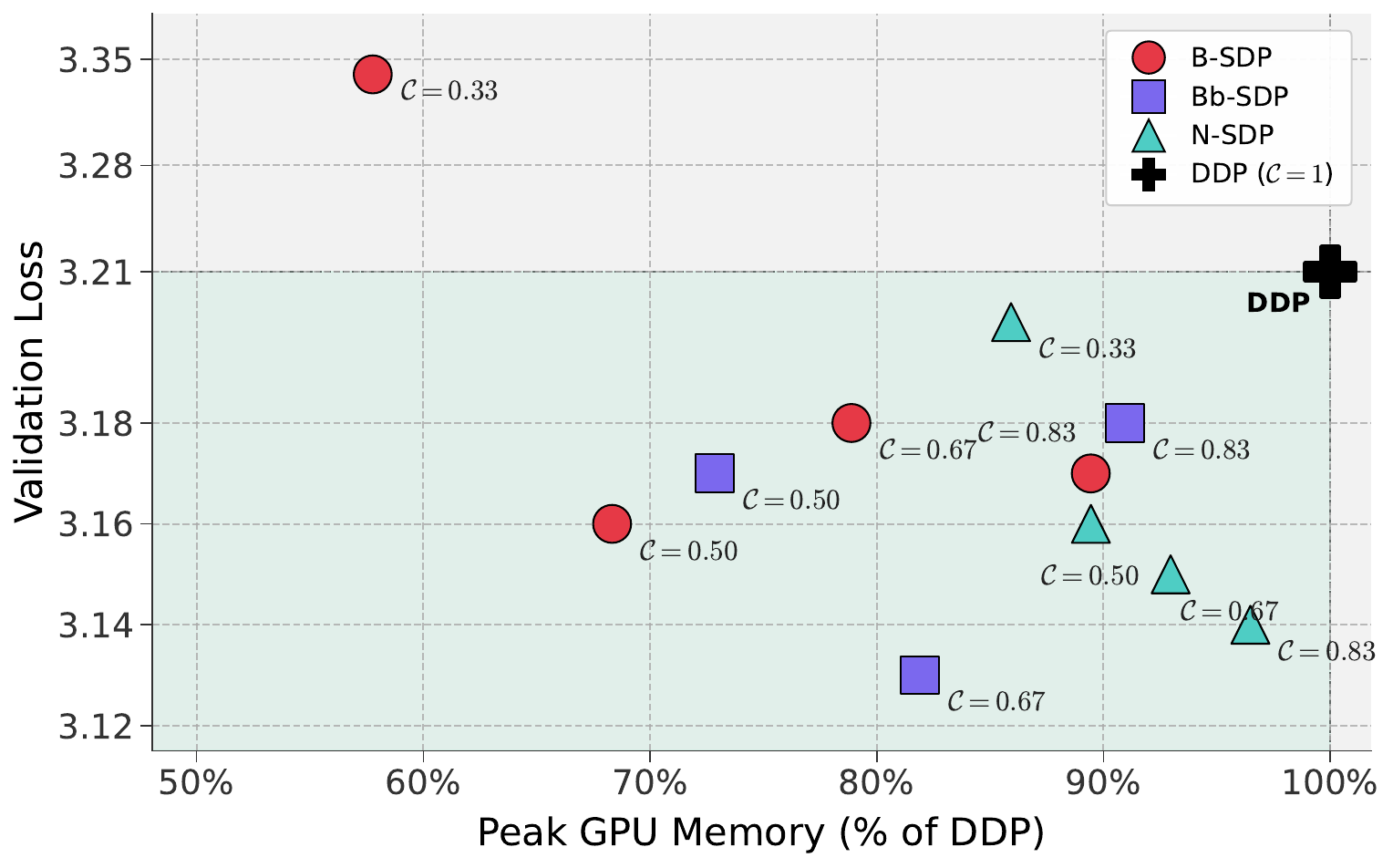}
  \caption{\small \textbf{134M LLaMA: SDP Pareto-dominates DDP.}
  Each point is a 134M LLaMA trained under a FLOP-matched protocol; only the coverage ratio \cor\ (annotated) varies. The shaded \emph{Pareto region} marks configurations that strictly dominate DDP: lower peak GPU memory at no worse validation loss. Multiple SDP variants (\bsdp, \bbsdp, \nsdp) sit firmly inside this region. Comparison to the 500M and 1B results in main-text Figure~\ref{fig:pareto_hero}.}
  \label{fig:pareto_134M}
\end{figure}

\textbf{Learning-rate sweeps.} At both 500M and 1B, we ran a per-(method, coverage) learning-rate sweep over the grid $\{1\mathrm{e}{-3},\, 2\mathrm{e}{-3},\, 4\mathrm{e}{-3},\, 6\mathrm{e}{-3},\, 8\mathrm{e}{-3},\, 1\mathrm{e}{-2}\}$ (6 LRs) for every SDP variant (\bsdp, \bbsdp, \nsdp) at every coverage ratio reported in the main text, reporting the best validation loss per cell. The LRs shown in \autoref{tab:llm_hparams} ($4\mathrm{e}{-3}$ at 500M/1B, $8\mathrm{e}{-3}$ at 134M) are the DDP optima from this sweep; most SDP settings also pick $4\mathrm{e}{-3}$ as best, with a few low-coverage points preferring $6\mathrm{e}{-3}$ (\bsdp\ and \bbsdp\ at \cor$\le 0.5$). At 134M we evaluated a smaller LR grid ($\{4\mathrm{e}{-3},\, 8\mathrm{e}{-3}\}$) since the DDP optimum was already at the top of the range.

\section{Communication and Step-Time Profiling}
\label{app:comm_profiling}
The numbers in Table~\ref{tab:comm_1B} are collected as follows. NCCL byte counts are measured by intercepting \texttt{torch.distributed} collectives. Per-op and per-step timings come from CUDA events with a 50-step warm-up followed by 100 measurement steps. FSDP compute/communication overlap is captured via NVIDIA Nsight NVTX markers.

\section{Results on 134M Model}
\label{app:134M}
As visualized by the Pareto front in Figure~\ref{fig:pareto_134M}, \textbf{every SDP configuration strictly Pareto-dominates the corresponding DDP baseline} achieve both lower validation loss and lower peak memory than DDP, confirming that subnetwork training is not merely a speed/accuracy tradeoff but a strict improvement on both axes.

\section{ResNet-18 SDP}
\label{app:resnet-18}

\begin{figure*}[t!]
    \centering
    \subfloat[Wide-ResNet-18 on CIFAR-10 \label{fig:inf:wa}]
    {\includegraphics[trim={4 0 0 0},clip, width=0.5\textwidth]{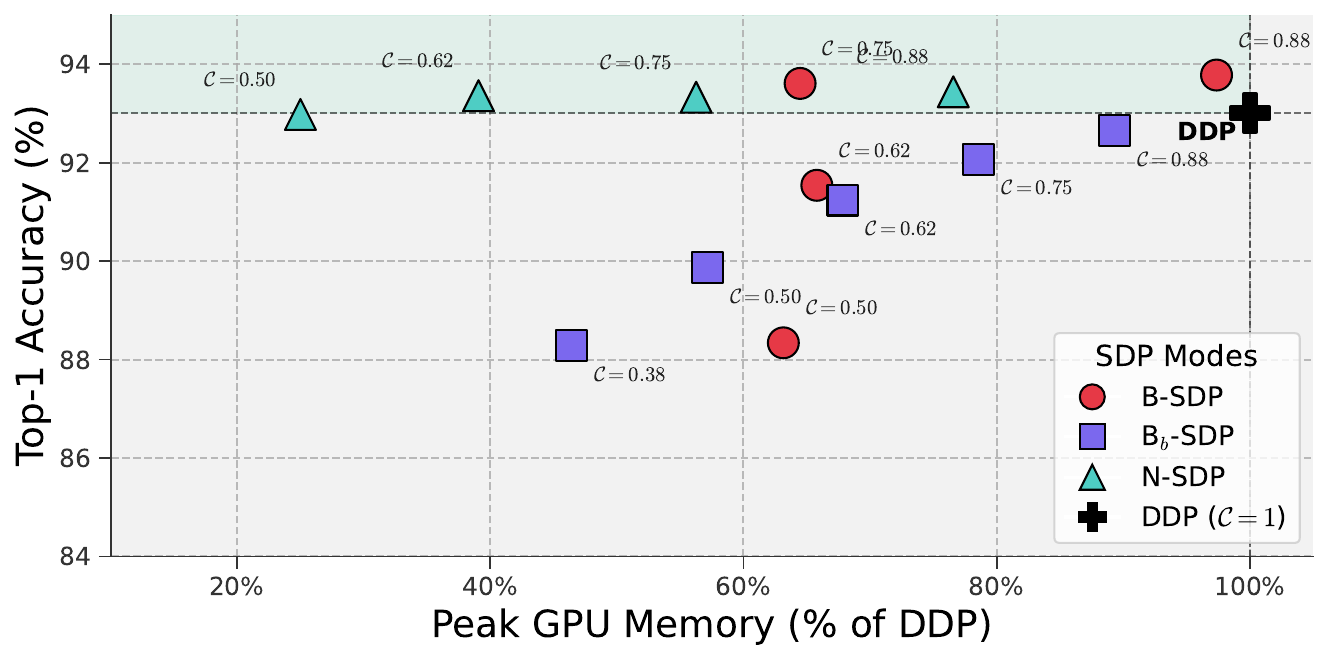}}
    \subfloat[Wide-ResNet-18 on CIFAR-100 \label{fig:inf:wb}]
    {\includegraphics[trim={4 0 0 0},clip, width=0.5\textwidth]{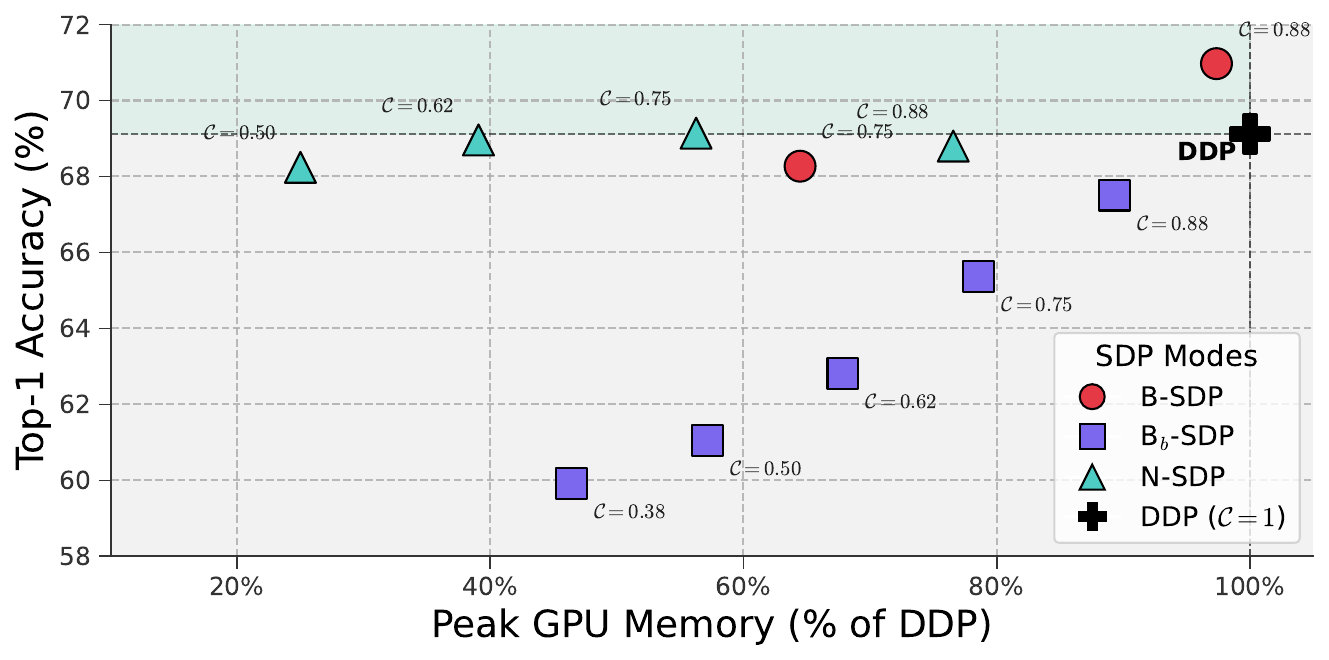}}

    \caption{\small Top-1 accuracy ($\uparrow$) versus fraction of DDP memory for \bsdp, \bbsdp, and \nsdp\ on WideResNet-18 across CIFAR-10 and CIFAR-100. The trends match ResNet-18 in main-text \autoref{fig:resnet18}: SDP reaches DDP-level accuracy at a fraction of the memory budget, and WRN-18 is more robust than RN-18 at high sparsity (e.g., \nsdp\ at \hlb{$\boldsymbol{25\%}$ of DDP memory} remains within the Pareto region on CIFAR-10). Per-configuration numbers are tabulated in \autoref{tab:rn_wrn_combined} and \autoref{tab:rn_wrn_linear}.}
    \label{fig:wrn18}
\end{figure*}

\begin{table*}[h]
\ifthenelse{\boolean{useratio}}{
\caption{\small Top-1 test accuracy (\%) ($\uparrow$) with \textbf{RN-18 and WRN-18} using a cosine annealing scheduler across different coverage ratios (\cor) comparing \nsdp, \bsdp, and \bbsdp\ with standard DDP (\cor$=1$). \colorbox{lightblue}{Blue cells} indicate configurations that match or exceed DDP accuracy. Notably, several configurations maintain this parity for coverage ratios \cor$\ge5/8$ and \cor$\ge4/8$ with RN-18 and WRN-18 respectively. Furthermore, at extreme sparsity {(\cor$=3/8$)} \bbsdp\ avoids performance collapse.}
}{
\caption{\small \textbf{ResNet-18 (RN-18) and WideResNet-18 (WRN-18) with cosine annealing.}
Top-1 test accuracy (\%) ($\uparrow$) across varying values of $p$ for Neuron-Level SDP (\nsdp), Block-Level SDP (\bsdp), and Block-Level Backward (\bbsdp). ($p=8$) denotes standard distributed data-parallel (DDP). Cells highlighted in \colorbox{lightblue}{light blue} are within the DDP result’s error bars or higher.}
}

\centering
\resizebox{0.9\linewidth}{!}{
\begin{tabular}{llcccccc}
\toprule
\multicolumn{8}{c}{\textbf{ResNet-18 (RN-18)}} \\
\midrule
\textbf{Dataset} & \textbf{Masking} &
\ifthenelse{\boolean{useratio}}{
\textbf{DDP (\cor $=1$)} &
\textbf{\cor $=7/8$} &
\textbf{\cor $=6/8$} &
\textbf{\cor $=5/8$} &
\textbf{\cor $=4/8$} &
\textbf{\cor $=3/8$} \\
}{
\textbf{DDP ($p=8$)} &
\textbf{$p=7$} &
\textbf{$p=6$} &
\textbf{$p=5$} &
\textbf{$p=4$} &
\textbf{$p=3$} \\
}
\midrule

\multirow{3}{*}{CIFAR-10}
  & \nsdp\ & \multirow{3}{*}{92.45 {\tiny$\pm$0.14}} &   \cellcolor{lightblue}{92.81 {\tiny$\pm$0.23}} &   \cellcolor{lightblue}{92.72 {\tiny$\pm$0.23}} &   \cellcolor{lightblue}{92.49 {\tiny$\pm$0.09}} &   {91.47 {\tiny$\pm$0.29}} & 22.56 {\tiny$\pm$2.04} \\
  & \bsdp\ &   &  \cellcolor{lightblue}{\textbf{93.18} {\tiny$\pm$0.16}} &  \cellcolor{lightblue}{92.89 {\tiny$\pm$0.18}} & 89.52 {\tiny$\pm$0.16} & 84.72 {\tiny$\pm$0.40} &   {42.68 {\tiny$\pm$2.09}} \\
  & \bbsdp &                & {92.14 {\tiny$\pm$0.14}} & 91.33 {\tiny$\pm$0.02} & 90.24 {\tiny$\pm$0.04} & 88.80 {\tiny$\pm$0.11} &    {87.91 {\tiny$\pm$0.29}} \\
\midrule

\multirow{3}{*}{CIFAR-100}
  & \nsdp\ & \multirow{3}{*}{68.62 {\tiny$\pm$0.01}} &   \cellcolor{lightblue}{69.02 {\tiny$\pm$0.14}} & \cellcolor{lightblue}{68.42 {\tiny$\pm$0.35}} &   {67.69 {\tiny$\pm$0.59}} &   {65.20 {\tiny$\pm$0.12}} &   {9.79 {\tiny$\pm$2.51}} \\
  & \bsdp\ &                          &  \cellcolor{lightblue}{\textbf{70.14} {\tiny$\pm$0.48}} &  \cellcolor{lightblue}{68.84 {\tiny$\pm$0.28}} & 54.27 {\tiny$\pm$0.51} & 36.20 {\tiny$\pm$0.01} & 7.03 {\tiny$\pm$0.40} \\
  & \bbsdp &                 & 67.33 {\tiny$\pm$0.43} & 64.90 {\tiny$\pm$0.24} & 61.87 {\tiny$\pm$0.16} &  {59.89 {\tiny$\pm$0.35}} &   {58.73 {\tiny$\pm$0.39}} \\
\midrule
\multicolumn{8}{c}{\textbf{WideResNet-18 (WRN-18)}} \\
\midrule
\textbf{Dataset} & \textbf{Masking} &
\ifthenelse{\boolean{useratio}}{
\textbf{DDP (\cor $=1$)} &
\textbf{\cor $=7/8$} &
\textbf{\cor $=6/8$} &
\textbf{\cor $=5/8$} &
\textbf{\cor $=4/8$} &
\textbf{\cor $=3/8$} \\
}{
\textbf{DDP ($p=8$)} &
\textbf{$p=7$} &
\textbf{$p=6$} &
\textbf{$p=5$} &
\textbf{$p=4$} &
\textbf{$p=3$} \\
}
\midrule
\multirow{3}{*}{CIFAR-10}
  & \nsdp\ & \multirow{3}{*}{{93.01 {\tiny$\pm$0.08}}} &   \cellcolor{lightblue}{93.44 {\tiny$\pm$0.03}} &   \cellcolor{lightblue}{93.33 {\tiny$\pm$0.04}} &  \cellcolor{lightblue}{93.36 {\tiny$\pm$0.21}} &   \cellcolor{lightblue}{92.98 {\tiny$\pm$0.09}} & 55.34 {\tiny$\pm$5.65} \\
  & \bsdp\ &            &  \cellcolor{lightblue}{\textbf{93.78} {\tiny$\pm$0.07}} &  \cellcolor{lightblue}{93.61 {\tiny$\pm$0.01}} & 91.54 {\tiny$\pm$0.16} & 88.34 {\tiny$\pm$0.15} &   {58.06 {\tiny$\pm$8.39}} \\
  & \bbsdp &                 & 92.65 {\tiny$\pm$0.14} & 92.07 {\tiny$\pm$0.10} &   {91.24 {\tiny$\pm$0.14}} &  {89.87 {\tiny$\pm$0.43}} &   {88.28 {\tiny$\pm$0.61}} \\
\midrule

\multirow{3}{*}{CIFAR-100}
  & \nsdp\ & \multirow{3}{*}{69.12 {\tiny$\pm$0.41}} &   \cellcolor{lightblue}{68.80 {\tiny$\pm$0.75}} &   \cellcolor{lightblue}{69.14 {\tiny$\pm$0.11}} &  \cellcolor{lightblue}{68.96 {\tiny$\pm$0.38}} &  \cellcolor{lightblue}{68.24 {\tiny$\pm$0.03}} & 44.74 {\tiny$\pm$0.47} \\
  & \bsdp\ &                          &  \cellcolor{lightblue}{\textbf{70.97} {\tiny$\pm$0.41}} & 68.27 {\tiny$\pm$0.12} & 56.90 {\tiny$\pm$0.31} & 42.23 {\tiny$\pm$0.85} & 9.25 {\tiny$\pm$0.53} \\
  & \bbsdp &                 & 67.51 {\tiny$\pm$0.28} &  {65.37 {\tiny$\pm$0.46}} &   {62.82 {\tiny$\pm$0.14}} &   {61.05 {\tiny$\pm$0.31}} &   {59.91 {\tiny$\pm$0.29}} \\
\bottomrule
\end{tabular}
}
\label{tab:rn_wrn_combined}
\vspace{-5pt}
\end{table*}

~\autoref{tab:rn_wrn_linear} presents the results comparing block masking and neuron masking when using a linear multi-step scheduler. We observe consistently superior performance with the \nsdp, especially at higher sparsity. For example, on CIFAR-100 with ResNet-18 and \nsdp\ at a coverage ratio of \ifthenelse{\boolean{useratio}}{\cor $=4/8$}{an overlap of $p = 4$}, the accuracy achieved with linear scheduling is 58.34\%, whereas \bsdp\ yields a significant degradation, reaching 40.30\%. Additionally, we find that the cosine scheduler delivers even higher performance at the same coverage for both 1x and 2x model sizes. These observations demonstrate that the effectiveness of the masking techniques is robust across different learning rate schedules and architectures, underscoring their scheduler-agnostic nature.

\begin{table*}[ht!]
\ifthenelse{\boolean{useratio}}{
\caption{\small Top-1 test accuracy (\%) with \textbf{RN-18 and WRN-18} using a multi-step linear scheduler across different coverage ratios (\cor) comparing \nsdp, \bsdp, and \bbsdp\ with standard DDP (\cor$=1$). \colorbox{lightblue}{Blue cells} indicate configurations that match or exceed DDP accuracy. Notably, several configurations maintain this parity for coverage ratios \cor$\ge5/8$ and \cor$\ge4/8$ with RN-18 and WRN-18 respectively. Furthermore, at extreme sparsity {(\cor$=3/8$)} \bbsdp\ avoids performance collapse.}
}{
\caption{\small \textbf{ResNet-18 (RN-18) and WideResNet-18 (WRN-18) with multi-step scheduler.}
Top-1 test accuracy (\%) across varying values of $p$ for Neuron-Level SDP (\nsdp), Block-Level SDP (\bsdp), and Block-Level Backward (\bbsdp). ($p=8$) denotes standard distributed data-parallel (DDP). Cells highlighted in \colorbox{lightblue}{light blue} are within the DDP result’s error bars or higher.}
}

\centering
\resizebox{0.9\textwidth}{!}{
\begin{tabular}{llcccccc}
\toprule
\multicolumn{8}{c}{\textbf{ResNet-18 (RN-18)}} \\
\midrule
\textbf{Dataset} & \textbf{Masking} &
\ifthenelse{\boolean{useratio}}{
\textbf{DDP (\cor $=1$)} &
\textbf{\cor $=7/8$} &
\textbf{\cor $=6/8$} &
\textbf{\cor $=5/8$} &
\textbf{\cor $=4/8$} &
\textbf{\cor $=3/8$} \\
}{
\textbf{DDP ($p=8$)} &
\textbf{$p=7$} &
\textbf{$p=6$} &
\textbf{$p=5$} &
\textbf{$p=4$} &
\textbf{$p=3$} \\
}
\midrule

\multirow{3}{*}{CIFAR-10} & \nsdp &  \multirow{3}{*}{92.41 {\tiny$\pm$0.08}} & \cellcolor{lightblue}{{93.14 {\tiny$\pm$0.28}}}   &  \cellcolor{lightblue}{92.95 {\tiny$\pm$0.24}} &   \cellcolor{lightblue}{92.23 {\tiny$\pm$0.24}} &   {91.25 {\tiny$\pm$0.26}} &   {80.93 {\tiny$\pm$2.87}} \\
         & \bsdp &  &  \cellcolor{lightblue}{\textbf{93.18} {\tiny$\pm$0.13}} &   \cellcolor{lightblue}{92.64 {\tiny$\pm$0.32}} & 90.35 {\tiny$\pm$0.13} & 84.01 {\tiny$\pm$0.95} & 39.35 {\tiny$\pm$0.55} \\
         & \bbsdp & & 91.62 {\tiny$\pm$0.13} & 91.50 {\tiny$\pm$0.37} & 89.61 {\tiny$\pm$0.23} & 87.94 {\tiny$\pm$0.53} &  {82.18 {\tiny$\pm$0.43}} \\
\midrule

\multirow{3}{*}{CIFAR-100} & \nsdp &  \multirow{3}{*}{65.02 {\tiny$\pm$0.16}} &   \cellcolor{lightblue}{65.64 {\tiny$\pm$0.48}} &   \cellcolor{lightblue}{65.76 {\tiny$\pm$0.82}} &  \cellcolor{lightblue}{64.95 {\tiny$\pm$0.45}} &   {58.34 {\tiny$\pm$1.57}} &   {50.00 {\tiny$\pm$0.09}} \\
          & \bsdp &  &  \cellcolor{lightblue}{\textbf{67.56} {\tiny$\pm$0.47}} &  \cellcolor{lightblue}{65.81 {\tiny$\pm$0.14}} & 56.52 {\tiny$\pm$1.84} & 40.30 {\tiny$\pm$0.47} & 8.39 {\tiny$\pm$1.25} \\
          & \bbsdp & & 61.80 {\tiny$\pm$0.24} & 60.33 {\tiny$\pm$0.37} & 58.08 {\tiny$\pm$0.06} & 55.30 {\tiny$\pm$0.71} &  {55.16 {\tiny$\pm$1.03}} \\
\midrule
\multicolumn{8}{c}{\textbf{WideResNet-18 (WRN-18)}} \\
\midrule
\textbf{Dataset} & \textbf{Masking} &
\ifthenelse{\boolean{useratio}}{
\textbf{DDP (\cor $=1$)} &
\textbf{\cor $=7/8$} &
\textbf{\cor $=6/8$} &
\textbf{\cor $=5/8$} &
\textbf{\cor $=4/8$} &
\textbf{\cor $=3/8$} \\
}{
\textbf{DDP ($p=8$)} &
\textbf{$p=7$} &
\textbf{$p=6$} &
\textbf{$p=5$} &
\textbf{$p=4$} &
\textbf{$p=3$} \\
}
\midrule

\multirow{3}{*}{CIFAR-10} & \nsdp & \multirow{3}{*}{92.26 {\tiny$\pm$0.55}} &   \cellcolor{lightblue}{93.47 {\tiny$\pm$0.37}} &  \cellcolor{lightblue}{\textbf{93.97} {\tiny$\pm$0.10}} &  \cellcolor{lightblue}{93.74 {\tiny$\pm$0.14}} &  \cellcolor{lightblue}{92.51 {\tiny$\pm$0.07}} &   {88.23 {\tiny$\pm$2.19}} \\
         & \bsdp & &   \cellcolor{lightblue}{93.66 {\tiny$\pm$0.25}} &   \cellcolor{lightblue}{93.28 {\tiny$\pm$0.11}} & 91.19 {\tiny$\pm$0.24} & 86.90 {\tiny$\pm$0.99} & 41.11 {\tiny$\pm$1.03} \\
         & \bbsdp & & \cellcolor{lightblue}{91.95 {\tiny$\pm$0.93}} & \cellcolor{lightblue}{91.79 {\tiny$\pm$0.25}} & 90.04 {\tiny$\pm$0.19} & 89.20 {\tiny$\pm$0.11} &  {86.20 {\tiny$\pm$0.38}} \\
\midrule

\multirow{3}{*}{CIFAR-100} & \nsdp &  \multirow{3}{*}{69.19 {\tiny$\pm$0.09}} &  \cellcolor{lightblue}{\textbf{69.86} {\tiny$\pm$0.27}} &   \cellcolor{lightblue}{68.91 {\tiny$\pm$0.21}} &   {66.20 {\tiny$\pm$0.21}} &   {63.71 {\tiny$\pm$0.63}} &   {58.02 {\tiny$\pm$0.59}} \\
          & \bsdp & & \cellcolor{lightblue}{69.26 {\tiny$\pm$0.42}} & 68.04 {\tiny$\pm$0.10} & 59.44 {\tiny$\pm$1.01} & 44.82 {\tiny$\pm$1.56} & 6.94 {\tiny$\pm$0.95} \\
          & \bbsdp & & 66.93 {\tiny$\pm$0.12} & 64.44 {\tiny$\pm$0.10} & 62.27 {\tiny$\pm$0.59} & 58.52 {\tiny$\pm$0.35} &  {55.51 {\tiny$\pm$0.20}} \\
\bottomrule
\end{tabular}
}
\vspace{0.15cm}

\label{tab:rn_wrn_linear}
\end{table*}

\section{Limitations} 
\label{app:limitations}
Our largest evaluation is 1B-parameter LLaMA pre-training. Architectural coverage is limited to dense LLaMA-style transformers and ResNet / WideResNet / Swin; mixture-of-experts, encoder-decoder, and multi-modal architectures are out of scope. Forward-masking variants (\nsdp, \bsdp) collapse at very aggressive sparsity (\cor\,$\le 3/8$ on RN-18 / WRN-18).

\section{Broader Impacts}
\label{app:broader_impacts}
This work targets the systems-level efficiency of distributed training. SDP reduces per-worker memory and communication for both LLMs and image classifiers, which lowers the hardware floor for training models of a given scale and lets practitioners train larger models on the same cluster. The expected impact is to widen access to large-scale training for groups without access to top-tier interconnects or high-memory accelerators, and to reduce the energy footprint of a fixed training run by trimming activation memory and gradient traffic.

We do not introduce new model capabilities, datasets, or deployable artifacts. SDP is a drop-in replacement for distributed data parallelism and inherits the societal profile of whatever model it is used to train; it does not, on its own, enable a class of application that was not previously possible. We are therefore not aware of any foreseeable negative societal consequences specific to this contribution.

\end{document}